\documentclass{article} 
\usepackage{lmrl2025_workshop,times}

\usepackage{amsmath,amsfonts,bm}

\def\eqref#1{equation~\ref{#1}}

\def\1{\bm{1}}

\DeclareMathAlphabet{\mathsfit}{\encodingdefault}{\sfdefault}{m}{sl}
\SetMathAlphabet{\mathsfit}{bold}{\encodingdefault}{\sfdefault}{bx}{n}

\usepackage{hyperref}
\usepackage{url}
\usepackage{graphicx}
\usepackage[export]{adjustbox}
\usepackage{multirow}
\usepackage{ragged2e}
\title{Benchmarking and optimizing organism \\ wide single-cell RNA alignment methods}

\author{Juan Javier Díaz-Mejía\thanks{Equal contribution}, Elias Williams\footnotemark[1], Octavian Focsa, \\
\textbf{Dylan Mendonca, Swechha Singh, Brendan Innes \& Sam Cooper} \\
Phenomic AI Inc. \\
Toronto, ON M5T 1X5, Canada \\
\texttt{\{sam, javier, elias\}@phenomic.ai}
}

\usepackage[normalem]{ulem}

\lmrlfinalcopy 
\begin{document}

\graphicspath{ {figures/} }
\maketitle

\begin{abstract}

Many methods have been proposed for removing batch effects and aligning single-cell RNA (scRNA) datasets. However, performance is typically evaluated based on multiple parameters and few datasets, creating challenges in assessing which method is best for aligning data at scale. Here, we introduce the K-Neighbors Intersection (KNI) score, a single score that both penalizes batch effects and measures accuracy at cross-dataset cell-type label prediction alongside carefully curated small (scMARK) and large (scREF) benchmarks comprising 11 and 46 human scRNA studies respectively, where we have standardized author labels. Using the KNI score, we evaluate and optimize approaches for cross-dataset single-cell RNA integration. We introduce Batch Adversarial single-cell Variational Inference (BA-scVI), as a new variant of scVI that uses adversarial training to penalize batch-effects in the encoder and decoder, and show this approach outperforms other methods. In the resulting aligned space, we find that the granularity of cell-type groupings is conserved, supporting the notion that whole-organism cell-type maps can be created by a single model without loss of information.

\end{abstract}

\section{Introduction}

To build comprehensive organism-wide and inter-species maps of cell types and states, we must build integrated transcriptional atlases that combine studies and patient populations at scale \cite{Regev2017-dz}. The now large number of published scRNA studies creates an opportunity for building a largely aligned scRNA atlas that would enable standardized reference-based analysis and cross-dataset comparison \cite{Lotfollahi2024-vb}. However, the challenge in combining data from disparate scRNA studies remains due to batch effects \cite{Lotfollahi2024-vb, Lahnemann2020-kj, Gavish2023-bn}.

\newblock
While studies have looked at the alignment of batches within datasets or between a handful of datasets focused on a specific tissue type, few have looked at model alignment across studies from different tissue types and instruments, as would be required for the generation of a reference atlas. Meanwhile, those studies that have used models to align datasets across tissue types and studies have used supervised models trained on cell-type labels, such as scBERT, Celltypist, SCimilarity, and SATURN \cite{Yang2022-nr,Dominguez-Conde2022-hg,Heimberg2024-ri,Rosen2024-tq}. However, unsupervised alignment is often preferred for cell-type discovery and comparative analysis of cell-type variation \cite{Vasighizaker2022-ju}. Thus, there remains a demand for unsupervised methods that can verifiably align scRNA data at scale \cite{Lotfollahi2024-vb}.

\newblock
In this study, we propose a single-metric the K-Neighbors Intersection (KNI) score that combines the K-bet score \cite{Buttner2019-eu}, with accuracy at author label prediction. We use this score to evaluate and optimize the ability of models to align a small (MNIST-like) and large benchmark (ImageNet-like) dataset that we present with standardized author labels. On this benchmark, we introduce a variant of the popular single-cell Variational Inference (scVI) model \cite{Lopez2018-cs}, Batch Adversarially trained single-cell Variational Inference (BA-scVI), that outperforms other approaches, including newer foundational models. The resulting embedding space maintains the cell-type clusters identified in the studies that comprise the benchmark, supporting the notion that a single unsupervised model can be used for organism-wide scRNA integration. Moreover, the prediction of cell types outside of the benchmark correlates well with benchmark scores, indicating author labels are a valid approximation of ground truth.

\section{Related Work}

\subsection{Metrics and benchmarks for assessment of scRNA alignment quality} 

The most comprehensive assessment of metrics and methods for scRNA alignment performance to date is performed in \cite{Luecken2020-bl}. Here, the authors benchmark 16 commonly used methods in 13 integration tasks. Performance on each task is assessed by a score that integrates five batch correction metrics, with nine metrics focused on the conservation of biological signal. Notably, this assessment includes the kBET score for batch effects \cite{Buttner2019-eu}, and F1 accuracy metric at cell-type prediction. This study has proven extremely valuable in helping authors pick specific models for specific scRNA analysis use cases \cite{Luecken2020-bl}. However, integrating these metrics after averaging of all data points creates a lower bar for success. Namely, a data point can improve the score obtained by being labeled correctly \textbf{or} by being well integrated with other datasets. We propose that top-performing alignment methods must both integrate data points with other studies \textbf{and} enable accurate prediction of cell type. This requires a metric that integrates batch-effect detection with accuracy for individual data points.

\subsection{Supervised methods for cross-dataset cell-type prediction} Recent studies have looked to use cross-dataset prediction of cell-type labels as a means of automated annotation. Notably, Fischer et al. describe a deep-learning architecture scTAB that performs well at cross-dataset cell-type prediction on datasets generated from 10X technologies \cite{Fischer2024-jk}. Here, the authors leverage the CELLxGENE corpus and schema, with automated filtering of cell-type groups for those having more fine-grained cell-type labels, with over 5000 instances, in at least 30 donors. They benchmark existing approaches and determine that the scTAB architecture performs best at cell type prediction with donor versus study level holdout of cells \cite{Fischer2024-jk}. Ergen et al. also recently described a consensus-based approach to cell-type annotation \cite{Ergen2024-fx}. Here, the authors develop a tool that performs unsupervised / semi-supervised alignment of scRNA datasets using three common approaches, BBKN, Scanorama, scVI, and scANVI, prior to the application of eight supervised models and then selection of the consensus cell-type label from these methods. This meta-approach performs well at unsupervised cell-type classification on a lung-tissue atlas but highlights that no single method has yet emerged that solves the task of new cell-type annotation. Our work here uses cell-type labels to assess performance at unsupervised alignment \textbf{versus} using cell-type labels at train time to maximize prediction accuracy. We propose that the top-performing unsupervised methods are likely better for cell-type, subtype, and state discovery as they are less susceptible to over-fitting. We also hypothesize that a single top-performing model can be identified that supersedes other approaches versus building off of a consensus.

\section{Results}
\subsection{The KNI score is a single-metric that evaluates scRNA alignment quality}

In scRNA dataset alignment, the task is to reduce gene space for individual datasets into a shared low-dimensional representation of cell types that are aligned between studies. It is important that this space captures underlying biological variation versus noise introduced by the experiment (batch-effects). Previously, the kBET score has been developed as an effective means of testing for the presence of batch effects \cite{Buttner2019-eu}. Accuracy at cross-dataset cell-type prediction of author labels also represents a gold-standard metric for evaluating the preservation of biological signal \cite{Dominguez_Conde2021-ne}. We propose the K-Neighbors Intersection (KNI) score as a score that combines these metrics at the level of individual data points Figure \ref{fig:fig1}a. To calculate the KNI we consider the set $C=\{c_1,c_2...c_n\}$ of cells in a low-dimensional cell-type feature space where each cell $c_i$ is defined by its coordinates $x_i$, batch identifier $b_i$, and cell-type identifier $t_i$. The distance function D between two cells is the Euclidean distance between their embedded coordinates. For the KNI, we thus identify the k-nearest neighbors for each cell $c_k$ as per a K-nearest neighbors search $K=\{c_i: D(c_k,c_i )\leq D(c_k,c_j )\ for\ all\ j \neq k\ and\ |K| = k\}$. For each cell $c_k$, we then identify a subset $B$ of $K$ in which cells have the same batch identifiers, defined as $B= \{c_i \in K:b_k = b_i \}$. Each cell $c_k$ is then labeled as either (1) an outlier if the number of elements in B is above a cutoff number $\tau < k$, i.e., too many nearest neighbors belong to the same batch (the K-bet score); or (2) the most common label from cells in $K$ but not $B$ (cross-dataset prediction accuracy).

$$ L(c_k) = 
\begin{Bmatrix}
null \  & |B| \geq \tau\\
mode(t_i : c_i \in K-B) & |B| < \tau
\end{Bmatrix} 
$$

The KNI score is then the total number of predicted labels that match author-standardized labels:
$$\text{KNI} = \frac{1}{n}\sum_{k=1}^n \mathbf{1} [L(c_k) = t_k]$$

We term this metric the K-nearest Neighbor Intersection (KNI) score since it evaluates accuracy at the intersection of batches (Appendix \ref{app1}). We find this metric has desirable properties when compared with other metrics on a simple theoretical test case (Appendix \ref{app1}), as well as a real-world scRNA dataset with simulated batch effects and noise (Appendix \ref{app2}). We also find that a Radius-based search can be used in place of the K-neighbors search such that,
$K=\{c_i: D(c_k,c_i ) \leq r\ for\ all\ i \neq k\}$. In the resulting score that we term the RbNI, a threshold percent of ‘self’ data points $\tau*$ is also used; in contrast to the KNI, cells with no neighbors within the radius $r$ are also given an outlier label such that, 

$$ L(c_k) = 
\begin{Bmatrix}
null \ & |K| = 0 \\
null \  & |B| \geq \tau*\\
mode(t_i : c_i \in K-B) & |B| < \tau*
\end{Bmatrix} 
$$

While we focus on the KNI, we also report key results under the RbNI metric and find it has similar properties to the KNI (Appendix \ref{app1},\ref{app2}), indicating neighborhood search strategy is not a key factor.

\subsection{Evaluation of models using the KNI score on scMARK}

We first used the KNI to evaluate aligned cell-type spaces generated by alignment approaches on a small MNIST-like benchmark to focus on scaling top-performing methods. For this, we curated the scMARK, a benchmark comprising of 11 high-quality scRNA publications from different labs, with a 10,000 random cell sample from each study (see Appendix \ref{app3} for details). We recorded 29 standardized cell-type labels occurring in two or more studies and 13,865 genes in common across the 11 studies. Of the 11 studies, 10 studies were produced using 10X Chromium technology, while \cite{Azizi2018-wb} was generated using the inDrop. We selected the 11 studies such that each cell-type/tissue-type combination appears in at least two studies, and thus, no ‘true’ outlier cell types should exist.

\begin{figure}[h]

\includegraphics[width=1.0\textwidth]{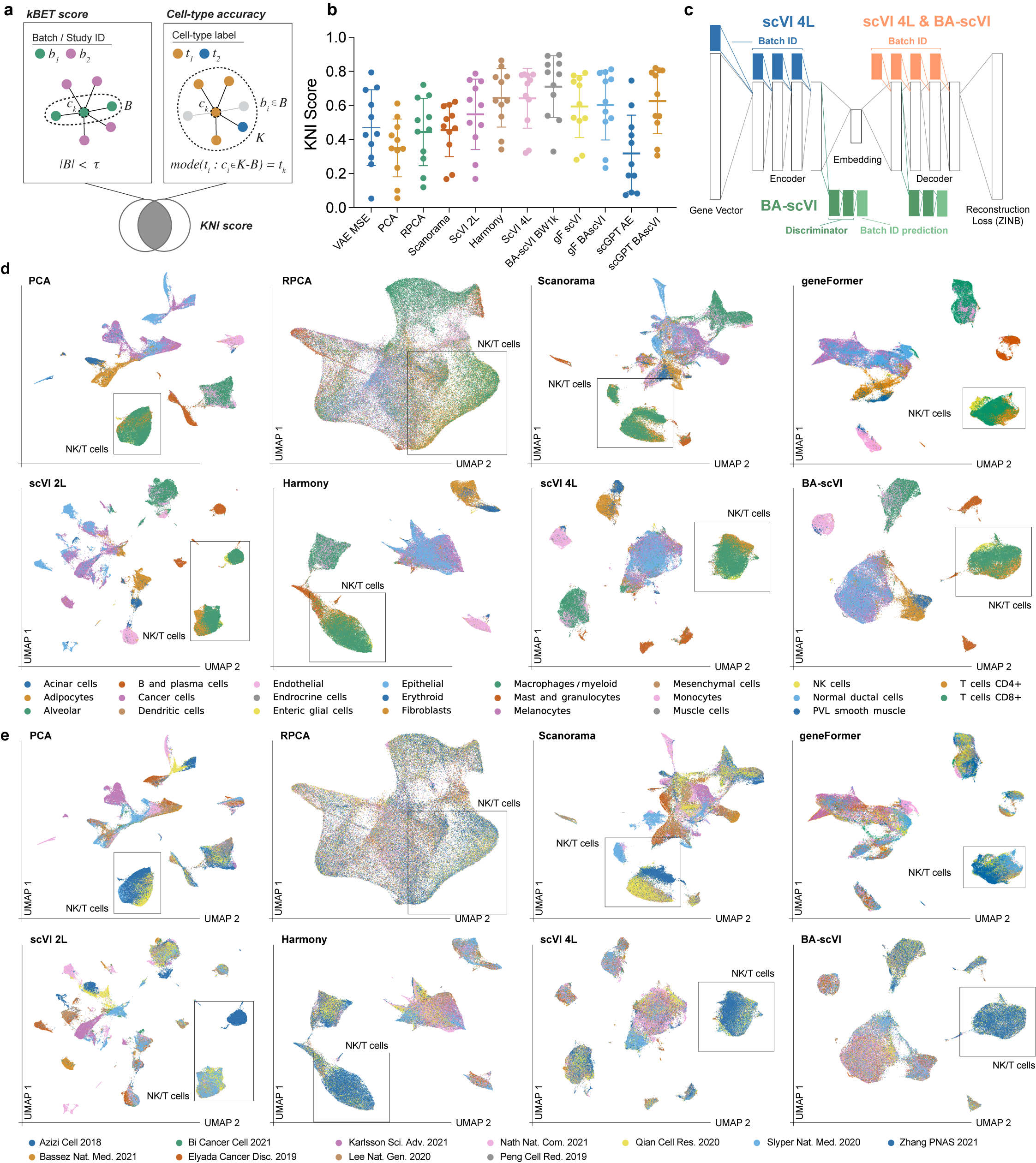}

\caption{Comparison of Model Performance on scMARK: a) The KNI score combines accuracy at cell-type labeling with batch-effect correction (KBET) at the data-point level; b) The KNI score is used to assess scRNA alignment model performance at aligning the 11 datasets in scMARK. KNI scores here are plotted for each of the 11 datasets alongside the dataset mean and standard deviation. A perfect score is 1; c) comparison of the scVI and BA-scVI architectures; d) UMAP projections of alignments produced by the eight different methods where cells are colored by author-provided ‘ground-truth’ cell-type label; e) UMAP projections as in (d), but colored by study. The NKT-cell grouping is highlighted to show variation in cell-type alignment quality between the methods notably the \cite{Azizi2018-wb} study (dark blue) is performed on the in-drop.}

\label{fig:fig1}
\end{figure}

\newblock
We used the KNI to evaluate the ability of commonly used methods to reduce scMARK into a shared 10-dimensional embedding space (Figure \ref{fig:fig1}b). Namely: (1) Principal Component Analysis (PCA) applied to highly variable genes (PCA)\cite{Kiselev2018-sa}; (2) Reciprocal PCA as described in the Seurat workflow (RPCA) \cite{Hao2021-zp}; (3) Scanorama \cite{Hie2019-kc}; (4) Variational Auto-Encoder with Mean Squared Error Loss (VAE MSE; included as a base-case for more advanced deep-learning tools); (5) Harmony, \cite{Korsunsky2019-lf}; (6) Base-line Single-cell Variational Inference (scVI)\cite{Lopez2018-cs}; (7) An optimized scVI, where we picked a scVI architecture that scores well under the KNI (Appendix 4); (8) A variant of the scVI architecture that we introduce here and that leverages an adversarial training to further enforce batch-effect removal as has been proposed e.g., in \cite{Shaham2018-kk} or \cite{Cao2020-fl}. We term this model Batch-Adversarial scVI (BA-scVI). In this model a discriminator is trained to predict batches on layers adjacent to the central embedding layer, versus the embedding layers itself, as we find this improves stability in training (Figure \ref{fig:fig1}c). Secondly, we removed batch-ID concatenation from the encoder (as is performed in scVI) to enable inference without batch-ID ; (9) geneFormer \cite{Theodoris2023-qz}, and (10) scGPT \cite{Cui2024-oc} – where in both cases we fine-tune an auto-encoder to reduce the transformed gene-space into a shared space as per recommendation;  and finally (11) geneFormer and (12) scGPT – where we use BA-scVI to reduce the transformed gene space into a 10-dimensional shared space (parameters for these models and details on the BA-scVI architecture are given in Appendix \ref{app4}).

\begin{table}
\begin{tabular}{ |p{3cm}||p{2.2cm}|p{2.2cm}|p{2cm}|p{2cm}|}
 \hline
 \multicolumn{5}{|c|}{Table \ref{table:1}: KNI and RbNI scores on the scMARK and scREF benchmarks} \\
 \hline
Model/Method & scMARK KNI & scMARK RbNI & scREF KNI & scREF RbNI\\
 \hline
 VAE MSE            & 0.352  & 0.322 &    -    &  - \\
 PCA                & 0.470  & 0.004 &  0.483 &  0.5123 \\
 RPCA               & 0.456  & 0.456 &    -    &  - \\
 Scanorama          & 0.446  & 0.477 &    -    &  - \\
 scVI 2L            & 0.550  & 0.592 &    -    &  - \\
 Harmony            & 0.646  & 0.615 &  0.488 &  0.4862 \\
 scVI 4L            & 0.643  & 0.645 &  0.578 &  0.586 \\
 BA-scVI            & \textbf{0.711}  & \textbf{0.687} &  \textbf{0.632} &  \textbf{0.619} \\
 gF scVI    & 0.596  & 0.637 &  0.399 &  0.437 \\
 gF BA-scVI & 0.604  & 0.641 &  0.400 &  0.441 \\
 scGPT AE           & 0.319  & 0.459 &  0.468 &  0.478 \\
 scGPT BA-scVI      & 0.627  & 0.658 &  0.425 &  0.463\\

\hline
\end{tabular}

\label{table:1}
\end{table}

\newblock
Based on this framework, BA-scVI outperforms other methods at aligning the scMARK dataset under the KNI score and related RbNI score, Table \ref{table:1}). We found that the transformer models, geneFormer and scGPT, generated poor-quality alignments when using the suggested fine-tuning approach. Fine-tuning with BA-scVI improved results significantly but failed to improve the alignment quality generated from untransformed data. Qualitatively, in UMAP projections, BA-scVI resolved cell subtypes, e.g., CD4+ vs. CD8+ T-cells on challenging datasets, e.g., \cite{Azizi2018-wb}, while removing any clear batch groupings. More broadly, the quality of alignment and batch-effect removal aligned well with the KNI score as visualized by UMAP, the tool most commonly used by biologists in exploring this data type (Figure \ref{fig:fig1}d); this further supports the KNI’s value for alignment assessment. We also see that the KNI and RbNI scores align well, indicating that the score is independent of the neighbor search strategy.

\subsection{Evaluation of models using the KNI score on scREF}

\begin{figure}[h] 
\includegraphics[width=1.0\textwidth]{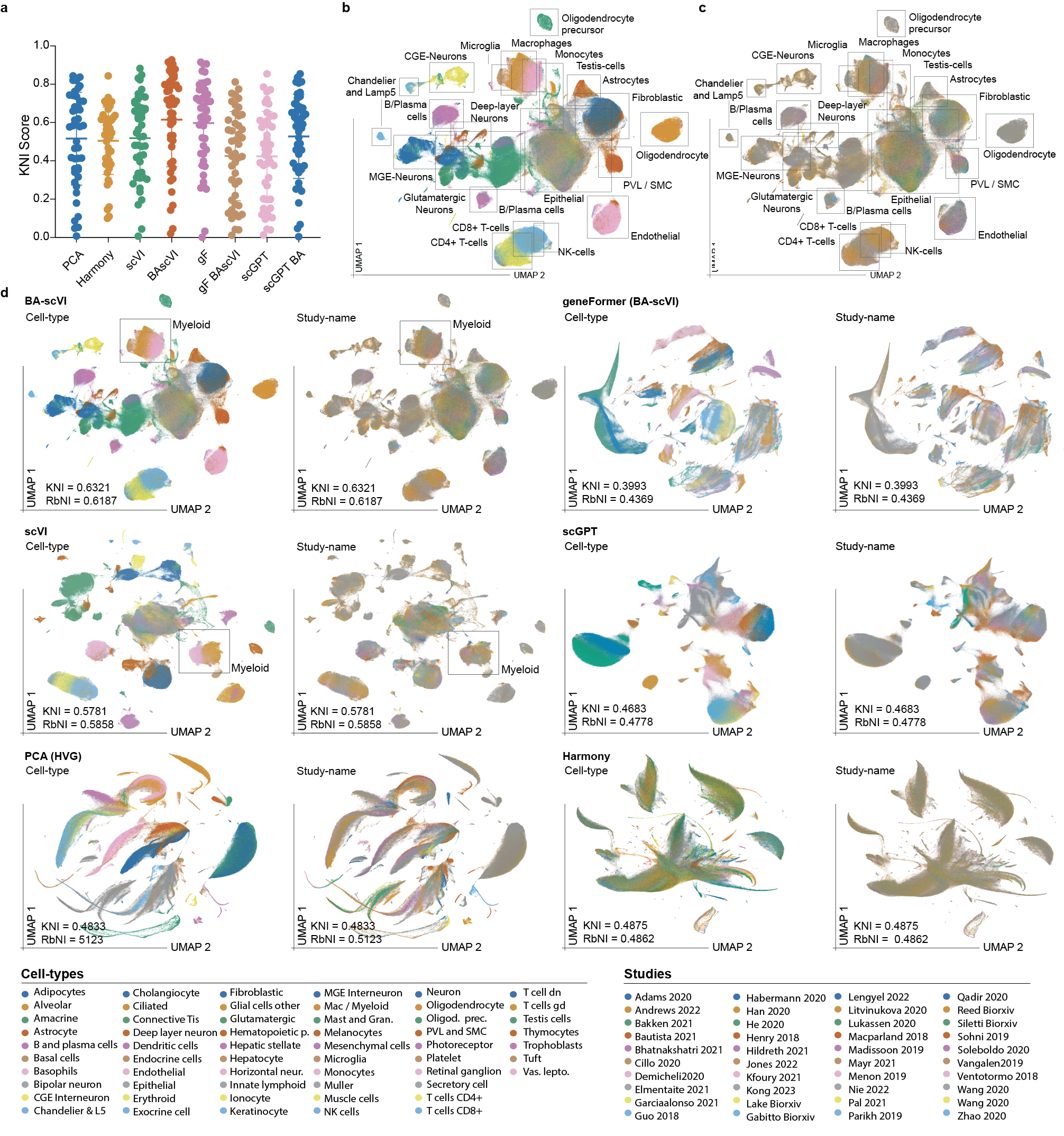}

\caption{Comparison of Model Performance on scREF: a)  KNI scores were determined for alignment of the 46 study scREF benchmark. Data points correspond to the average score achieved by the model on a study. The average score obtained on the entire benchmark plotted as a line; b) UMAP projections of the BA-scVI aligned scREF benchmark (n=1.27m), colored by ‘ground-truth’ standardized author cell-type label. The legend is omitted for brevity (coloring is the same as , boxes show major cell-type groupings; c) same projection as (b), colored by study name the legend is omitted; d) UMAP projections of scREF embedding spaces for the set of models presented colored by standardized author cell-type labels (left), and study (right).}
\label{fig:fig2}
\end{figure}

Next, we assessed how well top-performing methods can align scRNA data at an organism-wide scale. For this, we present the scREF benchmark, a collection of 46 human scRNA studies spanning 2,359 samples and 36 tissues, where for each dataset, quality checks have been performed and metadata standardized (see Appendix \ref{app3} for details). In scREF, we include organ-specific and human-wide datasets, e.g., the Tabula Sapiens \cite{Tabula_Sapiens_Consortium2022-bq} and the Human Cell Landscape \cite{Han2020-mx}. Importantly, scREF includes data from droplet-based (10X 5', 10X 3', 10X multiome, and Dropseq) and plate/bead based methods (Microwell-seq, Seq-Well and SMARTScribe) which allows for testing cross-technology alignment. Author-provided cell-type labels for 43 studies were acquired and standardized, while for three cases, we generated labels reproducing the original author's pipeline;  this resulted in 60 unique cell-type labels. Tissue-type labels were standardized for plotting and analyses, and a study-stratified sample was taken, seeking to balance tissue-type representation, leading to the final 1.21 million cell scREF benchmark.

\newblock
We then tested the top performing scRNA alignment models on this benchmark (Figure \ref{fig:fig2}a). Again, BA-scVI outperformed other approaches (Table \ref{table:1}). We also again saw that the transformer models, geneFormer and scGPT, introduced batch-effects that then could not be removed by fine-tuning either by author protocol or BA-scVI. Qualitatively, UMAP projections showed that BA-scVI produced a clear alignment by cell-type Figure \ref{fig:fig2}b. Notably, organism-wide studies from markedly different technologies Microwell-seq \cite{Han2020-mx}, and 10X \cite{Tabula_Sapiens_Consortium2022-bq} overlapped extensively with each other (Figure \ref{fig:fig2}c), indicating alignment independent of technology. We thus find that the BA-scVI model can be used to perform effective large-scale alignment. Again, the ability to distinguish cell types and the level of mixing observed qualitatively in UMAP projections mapped well to the KNI score (Figure \ref{fig:fig2}d). A high degree of alignment was again observed between the KNI and RbNI scores (Table \ref{table:1}).

\subsection{BA-scVI alignment of the scREF benchmark dataset maintains cell-type granularity}

\begin{figure}[h]

\includegraphics[width=1.0\textwidth]{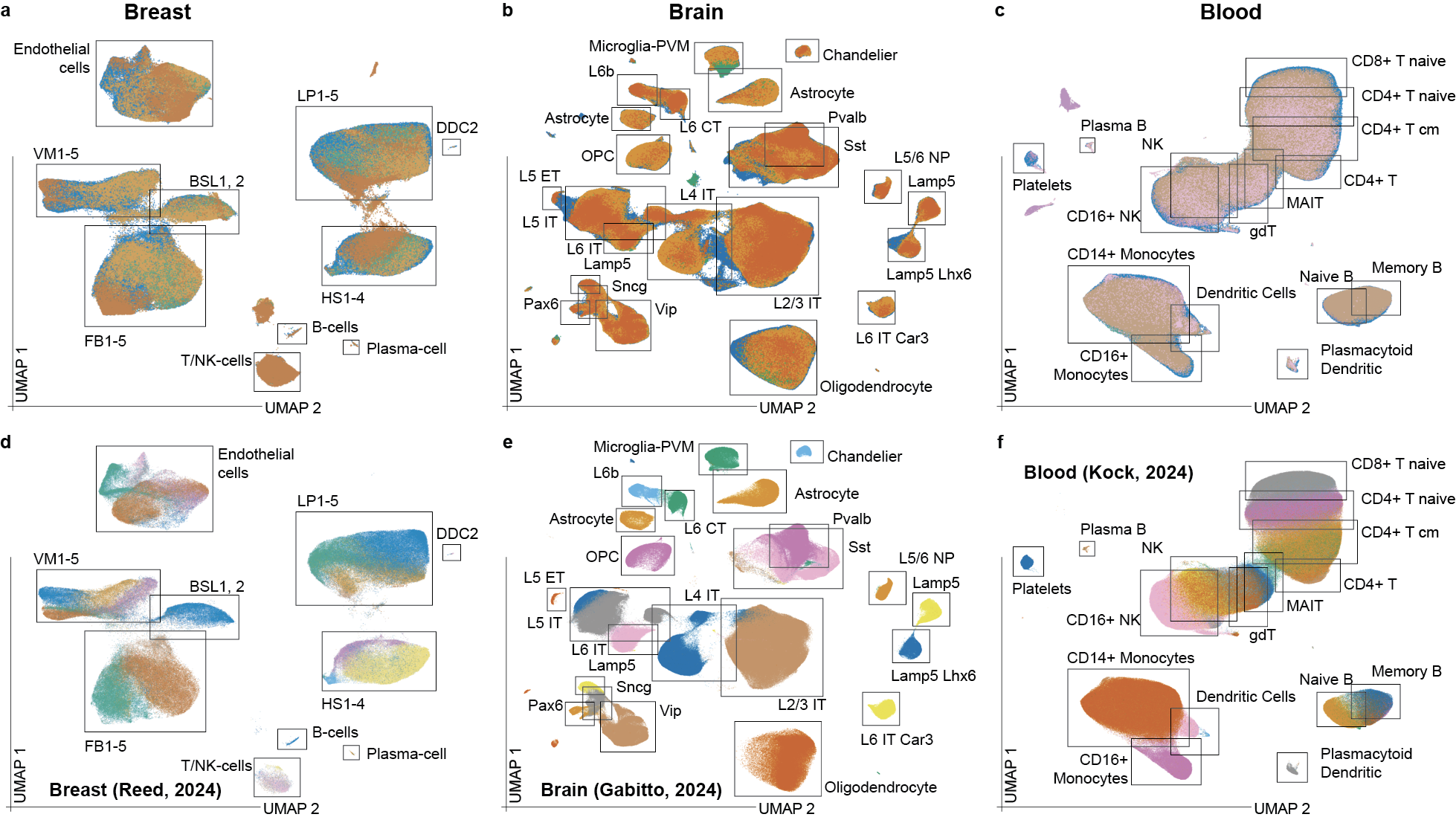}

\caption{BA-scVI scREF maintains cell-type granularity on alignment: a, b an c) 10-dimensional scRNA embeddings from BA-scVI corresponding to (a) Breast (n=0.4m cells), (b) Brain (n=4.8m cells), and (c) Blood (n=1.6m cells) tissue-types were projected into a 2-dimensional space with UMAP. Cells are colored by study name; d, e and f) The same UMAP projections but colored by original author labels for 3 example studies from each tissue type. Namely, (a) Breast \cite{Reed2024-ml} (n=0.3m cells), (b) Brain \cite{Gabitto2024-pm}(n=0.8m cells), and (c) Blood \cite{Kock2024-vi} (n=1m cells) The cell type and study legends omitted for brevity; major groupings are in boxes.}

\label{fig:fig3}
\end{figure}

A major concern in the atlas-building community is that aligning datasets reduces the granularity of cell-type detection. To qualitatively assess how well cell-type labeling are preserved at the organ level in the aligned cell-type space, we fit UMAP to the three best-represented tissues: breast (4 studies), brain (4 studies), and blood (7 studies). Supporting effective alignment with BA-scVI, we found significant overlap between studies (Figure \ref{fig:fig3}a-c). BA-scVI could also resolve ‘original author’ labels in UMAP projections of an example study for each tissue type, qualitatively supporting the preservation of cell-type resolution in the aligned space (Figure \ref{fig:fig3}d-f).

\newblock
Quantitatively, using a KNN accuracy test with 2-fold cross-validation, we find that the cell-type embeddings of the original author labels are conserved as a high degree of accuracy can be achieved on held-out data. Specifically, KNN accuracies of; (1) 83\% are obtained on a large breast dataset \cite{Reed2024-ml}; increasing to 96\% on ‘numerical’ subtype merging (e.g. cell subtypes ‘LP1’ to ‘LP5’ become ‘LP’); 2) 99\% for a brain study \cite{Gabitto2024-pm}; and 3) 83\% for the Kock et al. blood dataset where T-cell subtype label overlap is notably seen in projections in the original study \cite{Kock2024-vi}. Overall, this supports the preservation of cell-type granularity. 

\subsection{Comparison of the KNI score to KNN accuracy and the kBET score}

\begin{figure}[h]

\includegraphics[width=0.8\textwidth, center]{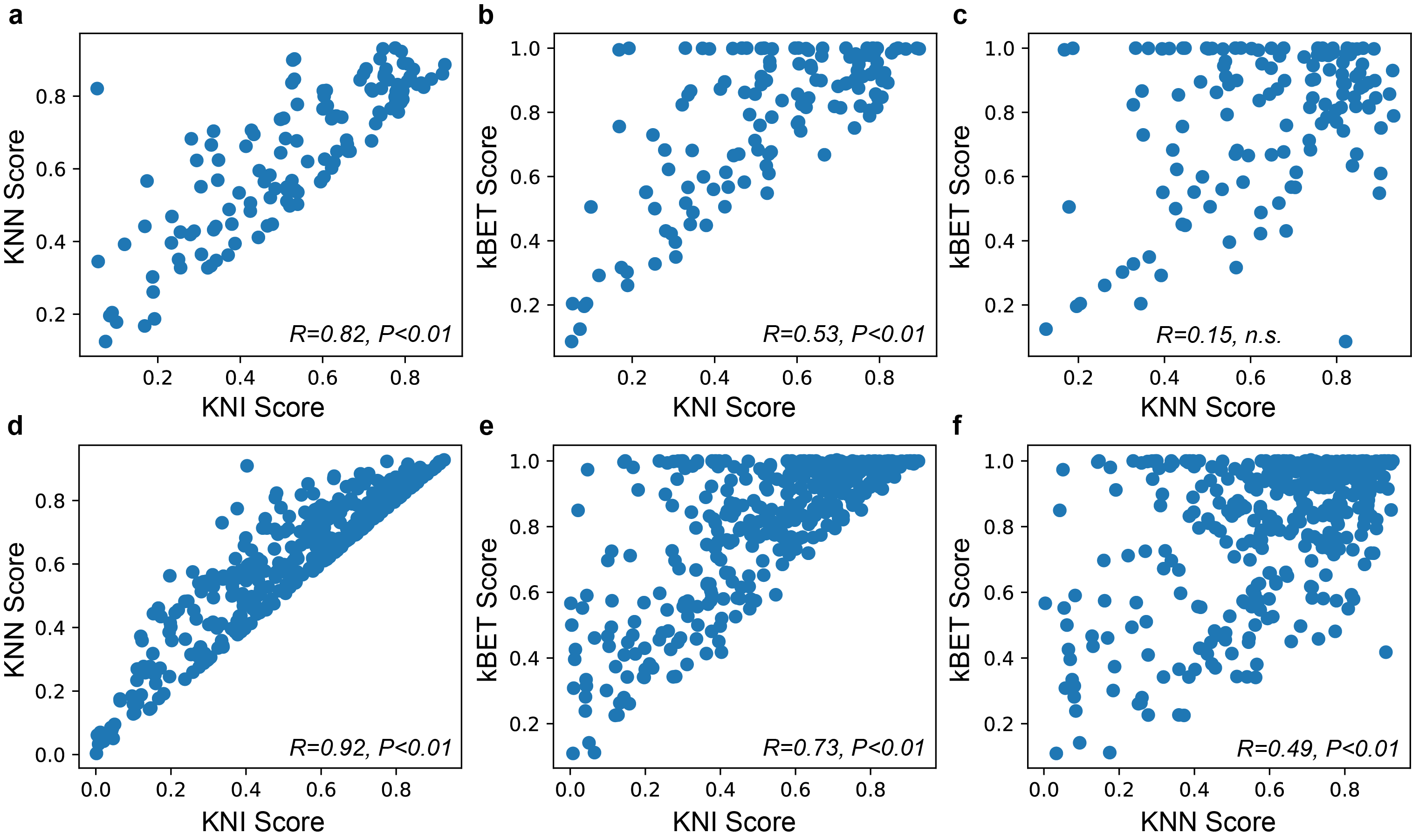}

\caption{Comparison of evaluation metrics on the scMARK and scREF benchmarks:  a) Correlation of KNI scores (x axis) and the KNN scores (y axis) on the studies in the scMARK benchmark achieved by the set of models tested in this paper. Here, the KNN classifier score is calculated using the nearest 50 neighbors from held-out datasets; b) Correlation of the KNI (x axis) score with the kBET score (y axis) \cite{Buttner2019-eu}; c) Correlation of the kBET score (y axis) with the KNN classifier score (x axis); d, e, and f) the same charts as (a, b, and c) but for models tested on studies in the scREF benchmark; to enable computation, here the KNN classifier score is calculated from the nearest 50 neighbors, less those from the same dataset.} 

\label{fig:fig4}
\end{figure}

We present the KNI score as a score that combines accuracy at predicting standardized author cell-type labels from held-out data with a KNN classifier, with batch-effect detection based on the kBET score \cite{Buttner2019-eu}. To better understand the relationship between these scores, we compared the performance each model achieves on each dataset under each of these three scores on scMARK and scREF. The results, including R-values and significance, are plotted in Figure \ref{fig:fig4}. This analysis highlights that the KNI score requires both a high KNN accuracy value and a high KBET score for a model to perform well on a specific dataset since the KNN-accuracy / kBET score is always equal to or better than the KNI score (Figure \ref{fig:fig4}a,b,d,e). When comparing the kBET and KNN scores with each other, we see much less correlation on both benchmarks (Figure \ref{fig:fig4}c,f). This means that, in many cases, approaches may be good at removing either batch effects or aligning cell types. In contrast, achieving both at the same time is a much higher bar. We also tested our ability to identify cell types not in the set of standardized cell-type labels described here (Appendix \ref{app5}) and found that KNI scores are comparable, supporting our assumption that standardized author labels can be used to approximate a ground truth. Overall, this supports the KNI score as a valuable new metric for evaluating scRNA alignment quality. 

\section{Discussion and Conclusions}
Most scRNA data alignment benchmarking studies have used only a handful of datasets to evaluate performance \cite{Pasquini2021-uk, Abdelaal2019-sr, Diaz-Mejia2019-qt}. This study presents scMARK and scREF as benchmark datasets for evaluating unsupervised model performance at scRNA alignment, at scale. This is in line with similar efforts underway, for example in image-based cell profiling \cite{Arevalo2024-kw}. A key concern with scRNA alignment metrics is highlighted in \cite{Wang2024-qc}. Specifically, the authors use a supervised model to demonstrate that cell 'islands' can form, whereby a model can group cells of the same cell-type label into a distinct island. By then forcing cell-type mixing within the island, very high scores can be achieved under the presented metrics. We consider this work to primarily show how easy it is to over-fit supervised models. Indeed, the authors highlight that weaker supervision is necessary \cite{Wang2024-qc}. The KNI score would be susceptible to thwarting by a supervised model; as such, we only present it as valuable for assessing unsupervised alignment (i.e., where only technical variables are used in training).

\newblock
We note that in our evaluation of the KNI metric, we have qualitatively compared UMAP projections to KNI scores. Although there are challenges with over-interpreting UMAP projections \cite{Chari2023-xl}, they remain the dominant approach used by biologists for the discovery and grouping of cell types. Thus, we consider it important that quantitative metrics align well with qualitative UMAP observations. We also stress that the KNI metric performs well in both theoretical and simulated scRNA alignment test cases (Appendix \ref{app1}, \ref{app2}), and is based on an intuitive combination of prior high-quality accuracy \cite{Dominguez_Conde2021-ne} and kBET metrics \cite{Buttner2019-eu}, providing firm quantitative support to its value.

\newblock
\cite{Lotfollahi2024-vb} highlight the need for models capable of reference-based alignment. Unlike the scVI architecture, BA-scVI does not require batch-ID for inference, thus we consider this model a good starting point for potential reference-based scRNA alignment tools. A key concern in the atlas-building community is the loss of granularity in cell-type resolution. The alignments we achieve with the BA-scVI model we present here provide compelling evidence that alignments can preserve cell-type granularity. While we encourage users to maintain caution, we think that with BA-scVI and further advances, this concern can be addressed, and the promise of reference-based scRNA analysis \cite{Lotfollahi2024-vb} realized.  

\subsubsection*{Meaningfulness Statement}
For us, a meaningful representation of life reflects life's natural evolution and development. Organisms share common ancestors and begin from a single-cell or branching event. We thus should be able to represent life with a continuous manifold, intersected with only rare discontinuities. In the space of single-cell biology, this means all cells, from all organisms, at any time, should be mappable to a single manifold; this free from technical effects and defined cell-type labels. We present significant results in unsupervised mapping of human scRNA data at scale, and hope this will advance efforts to map the wider manifold of life.

\subsubsection*{Reproducibility Statement }

All code and data (via download links with instructions) needed to reproduce the key results of this paper can be accessed at: \href{https://github.com/PhenomicAI/bascvi}{https://github.com/PhenomicAI/bascvi}.

\subsubsection*{Ethics Statement}
J. J. Diaz Mejia, E. Williams, S. Singh, O. Focsa, D. Mendonca, and B. Innes  are all equity holders of Phenomic AI Inc. and are/were employees of Phenomic AI Inc. S.C. is a founder, shareholder, and board-member of Phenomic AI Inc. Phenomic AI Inc. is a biotech developing new therapeutics targeting the tumor stroma.

\subsubsection*{Acknowledgments}

We would like to acknowledge Sean Grullon for help conceiving the project and Sarah Hackett for help curating scREF metadata. We would like to thank, Mike Briskin, Christopher Harvey, Girish Aakalu, Jimmy Ba, Oren Kraus, and Matthew Buechler for critical comments, revisions, and help on the manuscript. We appreciate the contributions of Allison Nixon, Ronen Schuster, Boris Hinz, in helping shape the paper and work. Finally, we acknowledge the rest of the team at Phenomic for feedback on the project in general.

\bibliography{lmrl2025_workshop}

\begin{thebibliography}{43}
\providecommand{\natexlab}[1]{#1}
\providecommand{\url}[1]{\texttt{#1}}
\expandafter\ifx\csname urlstyle\endcsname\relax
  \providecommand{\doi}[1]{doi: #1}\else
  \providecommand{\doi}{doi: \begingroup \urlstyle{rm}\Url}\fi

\bibitem[Abdelaal et~al.(2019)Abdelaal, Michielsen, Cats, Hoogduin, Mei, Reinders, and Mahfouz]{Abdelaal2019-sr}
Tamim Abdelaal, Lieke Michielsen, Davy Cats, Dylan Hoogduin, Hailiang Mei, Marcel J~T Reinders, and Ahmed Mahfouz.
\newblock A comparison of automatic cell identification methods for single-cell {RNA} sequencing data.
\newblock \emph{Genome Biol.}, 20\penalty0 (1):\penalty0 194, September 2019.

\bibitem[Arevalo et~al.(2024)Arevalo, Su, Ewald, van Dijk, Carpenter, and Singh]{Arevalo2024-kw}
John Arevalo, Ellen Su, Jessica~D Ewald, Robert van Dijk, Anne~E Carpenter, and Shantanu Singh.
\newblock Evaluating batch correction methods for image-based cell profiling.
\newblock \emph{Nat. Commun.}, 15\penalty0 (1):\penalty0 6516, August 2024.

\bibitem[Azizi et~al.(2018)Azizi, Carr, Plitas, Cornish, Konopacki, Prabhakaran, Nainys, Wu, Kiseliovas, Setty, Choi, Fromme, Dao, McKenney, Wasti, Kadaveru, Mazutis, Rudensky, and Pe'er]{Azizi2018-wb}
Elham Azizi, Ambrose~J Carr, George Plitas, Andrew~E Cornish, Catherine Konopacki, Sandhya Prabhakaran, Juozas Nainys, Kenmin Wu, Vaidotas Kiseliovas, Manu Setty, Kristy Choi, Rachel~M Fromme, Phuong Dao, Peter~T McKenney, Ruby~C Wasti, Krishna Kadaveru, Linas Mazutis, Alexander~Y Rudensky, and Dana Pe'er.
\newblock Single-cell map of diverse immune phenotypes in the breast tumor microenvironment.
\newblock \emph{Cell}, 174\penalty0 (5):\penalty0 1293--1308.e36, August 2018.

\bibitem[Bassez et~al.(2021)Bassez, Vos, Van~Dyck, Floris, Arijs, Desmedt, Boeckx, Vanden~Bempt, Nevelsteen, Lambein, Punie, Neven, Garg, Wildiers, Qian, Smeets, and Lambrechts]{Bassez2021-ya}
Ayse Bassez, Hanne Vos, Laurien Van~Dyck, Giuseppe Floris, Ingrid Arijs, Christine Desmedt, Bram Boeckx, Marlies Vanden~Bempt, Ines Nevelsteen, Kathleen Lambein, Kevin Punie, Patrick Neven, Abhishek~D Garg, Hans Wildiers, Junbin Qian, Ann Smeets, and Diether Lambrechts.
\newblock A single-cell map of intratumoral changes during anti-{PD1} treatment of patients with breast cancer.
\newblock \emph{Nat. Med.}, 27\penalty0 (5):\penalty0 820--832, May 2021.

\bibitem[Büttner et~al.(2019)Büttner, Miao, Wolf, Teichmann, and Theis]{Buttner2019-eu}
Maren Büttner, Zhichao Miao, F~Alexander Wolf, Sarah~A Teichmann, and Fabian~J Theis.
\newblock A test metric for assessing single-cell {RNA}-seq batch correction.
\newblock \emph{Nat. Methods}, 16\penalty0 (1):\penalty0 43--49, January 2019.

\bibitem[Caminschi et~al.(2008)Caminschi, Proietto, Ahmet, Kitsoulis, Shin~Teh, Lo, Rizzitelli, Wu, Vremec, van Dommelen, Campbell, Maraskovsky, Braley, Davey, Mottram, van~de Velde, Jensen, Lew, Wright, Heath, Shortman, and Lahoud]{Caminschi2008-ds}
Irina Caminschi, Anna~I Proietto, Fatma Ahmet, Susie Kitsoulis, Joo Shin~Teh, Jennifer C~Y Lo, Alexandra Rizzitelli, Li~Wu, David Vremec, Serani L~H van Dommelen, Ian~K Campbell, Eugene Maraskovsky, Hal Braley, Gayle~M Davey, Patricia Mottram, Nicholas van~de Velde, Kent Jensen, Andrew~M Lew, Mark~D Wright, William~R Heath, Ken Shortman, and Mireille~H Lahoud.
\newblock The dendritic cell subtype-restricted {C}-type lectin {Clec9A} is a target for vaccine enhancement.
\newblock \emph{Blood}, 112\penalty0 (8):\penalty0 3264--3273, October 2008.

\bibitem[Cao et~al.(2020)Cao, Wei, Lu, Yang, and Gao]{Cao2020-fl}
Zhi-Jie Cao, Lin Wei, Shen Lu, De-Chang Yang, and Ge~Gao.
\newblock Searching large-scale {scRNA}-seq databases via unbiased cell embedding with cell {BLAST}.
\newblock \emph{Nat. Commun.}, 11\penalty0 (1):\penalty0 3458, July 2020.

\bibitem[Chari \& Pachter(2023)Chari and Pachter]{Chari2023-xl}
Tara Chari and Lior Pachter.
\newblock The specious art of single-cell genomics.
\newblock \emph{PLoS Comput. Biol.}, 19\penalty0 (8):\penalty0 e1011288, August 2023.

\bibitem[Cui et~al.(2024)Cui, Wang, Maan, Pang, Luo, Duan, and Wang]{Cui2024-oc}
Haotian Cui, Chloe Wang, Hassaan Maan, Kuan Pang, Fengning Luo, Nan Duan, and Bo~Wang.
\newblock {scGPT}: toward building a foundation model for single-cell multi-omics using generative {AI}.
\newblock \emph{Nat. Methods}, February 2024.

\bibitem[Diaz-Mejia et~al.(2019)Diaz-Mejia, Javier Diaz-Mejia, Meng, Pico, MacParland, Ketela, Pugh, Bader, and Morris]{Diaz-Mejia2019-qt}
J~Javier Diaz-Mejia, J~Javier Diaz-Mejia, Elaine~C Meng, Alexander~R Pico, Sonya~A MacParland, Troy Ketela, Trevor~J Pugh, Gary~D Bader, and John~H Morris.
\newblock Evaluation of methods to assign cell type labels to cell clusters from single-cell {RNA}-sequencing data, 2019.

\bibitem[Domínguez~Conde et~al.(2021)Domínguez~Conde, Xu, Jarvis, Gomes, Howlett, Rainbow, Suchanek, King, Mamanova, Polanski, Huang, Fasouli, Mahbubani, Prete, Tuck, Richoz, Tuong, Campos, Mousa, Needham, Pritchard, Li, Elmentaite, Park, Menon, Bayraktar, James, Meyer, Clatworthy, Saeb-Parsy, Jones, and Teichmann]{Dominguez_Conde2021-ne}
C~Domínguez~Conde, C~Xu, L~B Jarvis, T~Gomes, S~K Howlett, D~B Rainbow, O~Suchanek, H~W King, L~Mamanova, K~Polanski, N~Huang, E~S Fasouli, K~T Mahbubani, M~Prete, L~Tuck, N~Richoz, Z~K Tuong, L~Campos, H~S Mousa, E~J Needham, S~Pritchard, T~Li, R~Elmentaite, J~Park, D~K Menon, O~A Bayraktar, L~K James, K~B Meyer, M~R Clatworthy, K~Saeb-Parsy, J~L Jones, and S~A Teichmann.
\newblock Cross-tissue immune cell analysis reveals tissue-specific adaptations and clonal architecture in humans.
\newblock \emph{bioRxiv}, pp.\  2021.04.28.441762, July 2021.

\bibitem[Domínguez~Conde et~al.(2022)Domínguez~Conde, Xu, Jarvis, Rainbow, Wells, Gomes, Howlett, Suchanek, Polanski, King, Mamanova, Huang, Szabo, Richardson, Bolt, Fasouli, Mahbubani, Prete, Tuck, Richoz, Tuong, Campos, Mousa, Needham, Pritchard, Li, Elmentaite, Park, Rahmani, Chen, Menon, Bayraktar, James, Meyer, Yosef, Clatworthy, Sims, Farber, Saeb-Parsy, Jones, and Teichmann]{Dominguez-Conde2022-hg}
C~Domínguez~Conde, C~Xu, L~B Jarvis, D~B Rainbow, S~B Wells, T~Gomes, S~K Howlett, O~Suchanek, K~Polanski, H~W King, L~Mamanova, N~Huang, P~A Szabo, L~Richardson, L~Bolt, E~S Fasouli, K~T Mahbubani, M~Prete, L~Tuck, N~Richoz, Z~K Tuong, L~Campos, H~S Mousa, E~J Needham, S~Pritchard, T~Li, R~Elmentaite, J~Park, E~Rahmani, D~Chen, D~K Menon, O~A Bayraktar, L~K James, K~B Meyer, N~Yosef, M~R Clatworthy, P~A Sims, D~L Farber, K~Saeb-Parsy, J~L Jones, and S~A Teichmann.
\newblock Cross-tissue immune cell analysis reveals tissue-specific features in humans.
\newblock \emph{Science}, 376\penalty0 (6594):\penalty0 eabl5197, May 2022.

\bibitem[Ergen et~al.(2024)Ergen, Xing, Xu, Kim, Jayasuriya, McGeever, Pisco, Streets, and Yosef]{Ergen2024-fx}
Can Ergen, Galen~K Xing, Chenling Xu, Martin Kim, Michael Jayasuriya, Erin McGeever, Angela~Oliveira Pisco, Aaron~M Streets, and N~Yosef.
\newblock Consensus prediction of cell type labels in single-cell data with {popV}.
\newblock \emph{Nat. Genet.}, 56:\penalty0 2731--2738, November 2024.

\bibitem[Fischer et~al.(2024)Fischer, Fischer, Mukhin, Isaev, Biederstedt, Villani, and Theis]{Fischer2024-jk}
Felix Fischer, David~S Fischer, Roman Mukhin, Andrey Isaev, Evan Biederstedt, Alexandra-Chloé Villani, and Fabian~J Theis.
\newblock {scTab}: Scaling cross-tissue single-cell annotation models.
\newblock \emph{Nat. Commun.}, 15\penalty0 (1):\penalty0 6611, August 2024.

\bibitem[Fontenot et~al.(2003)Fontenot, Gavin, and Rudensky]{Fontenot2003-og}
J~Fontenot, M~Gavin, and A~Rudensky.
\newblock {Foxp3} programs the development and function of {CD4+CD25+} regulatory {T} cells.
\newblock \emph{Nat. Immunol.}, 4:\penalty0 330--336, March 2003.

\bibitem[Gabitto et~al.(2024)Gabitto, Travaglini, Rachleff, Kaplan, Long, Ariza, Ding, Mahoney, Dee, Goldy, Melief, Agrawal, Kana, Zhen, Barlow, Brouner, Campos, Campos, Carr, Casper, Chakrabarty, Clark, Cool, Dalley, Darvas, Ding, Dolbeare, Egdorf, Esposito, Ferrer, Fleckenstein, Gala, Gary, Gelfand, Gloe, Guilford, Guzman, Hirschstein, Ho, Hupp, Jarsky, Johansen, Kalmbach, Keene, Khawand, Kilgore, Kirkland, Kunst, Lee, Leytze, Mac~Donald, Malone, Maltzer, Martin, McCue, McMillen, Mena, Meyerdierks, Meyers, Mollenkopf, Montine, Nolan, Nyhus, Olsen, Pacleb, Pagan, Peña, Pham, Pom, Postupna, Rimorin, Ruiz, Saldi, Schantz, Shapovalova, Sorensen, Staats, Sullivan, Sunkin, Thompson, Tieu, Ting, Torkelson, Tran, Valera~Cuevas, Walling-Bell, Wang, Waters, Wilson, Xiao, Haynor, Gatto, Jayadev, Mufti, Ng, Mukherjee, Crane, Latimer, Levi, Smith, Close, Miller, Hodge, Larson, Grabowski, Hawrylycz, Keene, and Lein]{Gabitto2024-pm}
Mariano~I Gabitto, Kyle~J Travaglini, Victoria~M Rachleff, Eitan~S Kaplan, Brian Long, Jeanelle Ariza, Yi~Ding, Joseph~T Mahoney, Nick Dee, Jeff Goldy, Erica~J Melief, Anamika Agrawal, Omar Kana, Xingjian Zhen, Samuel~T Barlow, Krissy Brouner, Jazmin Campos, John Campos, Ambrose~J Carr, Tamara Casper, Rushil Chakrabarty, Michael Clark, Jonah Cool, Rachel Dalley, Martin Darvas, Song-Lin Ding, Tim Dolbeare, Tom Egdorf, Luke Esposito, Rebecca Ferrer, Lynn~E Fleckenstein, Rohan Gala, Amanda Gary, Emily Gelfand, Jessica Gloe, Nathan Guilford, Junitta Guzman, Daniel Hirschstein, Windy Ho, Madison Hupp, Tim Jarsky, Nelson Johansen, Brian~E Kalmbach, Lisa~M Keene, Sarah Khawand, Mitchell~D Kilgore, Amanda Kirkland, Michael Kunst, Brian~R Lee, Mckaila Leytze, Christine~L Mac~Donald, Jocelin Malone, Zoe Maltzer, Naomi Martin, Rachel McCue, Delissa McMillen, Gonzalo Mena, Emma Meyerdierks, Kelly~P Meyers, Tyler Mollenkopf, Mark Montine, Amber~L Nolan, Julie~K Nyhus, Paul~A Olsen, Maiya Pacleb, Chelsea~M Pagan, Nicholas
  Peña, Trangthanh Pham, Christina~Alice Pom, Nadia Postupna, Christine Rimorin, Augustin Ruiz, Giuseppe~A Saldi, Aimee~M Schantz, Nadiya~V Shapovalova, Staci~A Sorensen, Brian Staats, Matt Sullivan, Susan~M Sunkin, Carol Thompson, Michael Tieu, Jonathan~T Ting, Amy Torkelson, Tracy Tran, Nasmil~J Valera~Cuevas, Sarah Walling-Bell, Ming-Qiang Wang, Jack Waters, Angela~M Wilson, Ming Xiao, David Haynor, Nicole~M Gatto, Suman Jayadev, Shoaib Mufti, Lydia Ng, Shubhabrata Mukherjee, Paul~K Crane, Caitlin~S Latimer, Boaz~P Levi, Kimberly~A Smith, Jennie~L Close, Jeremy~A Miller, Rebecca~D Hodge, Eric~B Larson, Thomas~J Grabowski, Michael Hawrylycz, C~Dirk Keene, and Ed~S Lein.
\newblock Integrated multimodal cell atlas of alzheimer's disease.
\newblock \emph{Nat. Neurosci.}, 27\penalty0 (12):\penalty0 2366--2383, December 2024.

\bibitem[Gavish et~al.(2023)Gavish, Tyler, Greenwald, Hoefflin, Simkin, Tschernichovsky, Galili~Darnell, Somech, Barbolin, Antman, Kovarsky, Barrett, Gonzalez~Castro, Halder, Chanoch-Myers, Laffy, Mints, Wider, Tal, Spitzer, Hara, Raitses-Gurevich, Stossel, Golan, Tirosh, Suvà, Puram, and Tirosh]{Gavish2023-bn}
Avishai Gavish, Michael Tyler, Alissa~C Greenwald, Rouven Hoefflin, Dor Simkin, Roi Tschernichovsky, Noam Galili~Darnell, Einav Somech, Chaya Barbolin, Tomer Antman, Daniel Kovarsky, Thomas Barrett, L~Nicolas Gonzalez~Castro, Debdatta Halder, Rony Chanoch-Myers, Julie Laffy, Michael Mints, Adi Wider, Rotem Tal, Avishay Spitzer, Toshiro Hara, Maria Raitses-Gurevich, Chani Stossel, Talia Golan, Amit Tirosh, Mario~L Suvà, Sidharth~V Puram, and Itay Tirosh.
\newblock Hallmarks of transcriptional intratumour heterogeneity across a thousand tumours.
\newblock \emph{Nature}, 618\penalty0 (7965):\penalty0 598--606, June 2023.

\bibitem[Han et~al.(2020)Han, Zhou, Fei, Sun, Wang, Chen, Chen, Wang, Tang, Ge, Zhou, Ye, Jiang, Wu, Xiao, Jia, Zhang, Ma, Zhang, Bai, Lai, Yu, Zhu, Lin, Gao, Wang, Wu, Zhang, Zhan, Zhu, Hu, Wang, Chen, Huang, Liang, Chen, Wang, Zhang, and Guo]{Han2020-mx}
Xiaoping Han, Ziming Zhou, Lijiang Fei, Huiyu Sun, Renying Wang, Yao Chen, Haide Chen, Jingjing Wang, Huanna Tang, Wenhao Ge, Yincong Zhou, Fang Ye, Mengmeng Jiang, Junqing Wu, Yanyu Xiao, Xiaoning Jia, Tingyue Zhang, Xiaojie Ma, Qi~Zhang, Xueli Bai, Shujing Lai, Chengxuan Yu, Lijun Zhu, Rui Lin, Yuchi Gao, Min Wang, Yiqing Wu, Jianming Zhang, Renya Zhan, Saiyong Zhu, Hailan Hu, Changchun Wang, Ming Chen, He~Huang, Tingbo Liang, Jianghua Chen, Weilin Wang, Dan Zhang, and Guoji Guo.
\newblock Construction of a human cell landscape at single-cell level.
\newblock \emph{Nature}, 581\penalty0 (7808):\penalty0 303--309, May 2020.

\bibitem[Hao et~al.(2021)Hao, Hao, Andersen-Nissen, Mauck, Zheng, Butler, Lee, Wilk, Darby, Zager, Hoffman, Stoeckius, Papalexi, Mimitou, Jain, Srivastava, Stuart, Fleming, Yeung, Rogers, McElrath, Blish, Gottardo, Smibert, and Satija]{Hao2021-zp}
Yuhan Hao, Stephanie Hao, Erica Andersen-Nissen, William~M Mauck, 3rd, Shiwei Zheng, Andrew Butler, Maddie~J Lee, Aaron~J Wilk, Charlotte Darby, Michael Zager, Paul Hoffman, Marlon Stoeckius, Efthymia Papalexi, Eleni~P Mimitou, Jaison Jain, Avi Srivastava, Tim Stuart, Lamar~M Fleming, Bertrand Yeung, Angela~J Rogers, Juliana~M McElrath, Catherine~A Blish, Raphael Gottardo, Peter Smibert, and Rahul Satija.
\newblock Integrated analysis of multimodal single-cell data.
\newblock \emph{Cell}, 184\penalty0 (13):\penalty0 3573--3587.e29, June 2021.

\bibitem[Heimberg et~al.(2024)Heimberg, Kuo, DePianto, Salem, Heigl, Diamant, Scalia, Biancalani, Turley, Rock, Bravo, Kaminker, Vander~Heiden, and Regev]{Heimberg2024-ri}
Graham Heimberg, Tony Kuo, D~DePianto, Omar Salem, Tobias Heigl, Nathaniel Diamant, Gabriele Scalia, Tommasso Biancalani, Shannon~J Turley, Jason~R Rock, Héctor~Corrada Bravo, Josh Kaminker, J~V Vander~Heiden, and Aviv Regev.
\newblock A cell atlas foundation model for scalable search of similar human cells.
\newblock \emph{Nature}, November 2024.

\bibitem[Hie et~al.(2019)Hie, Bryson, and Berger]{Hie2019-kc}
Brian Hie, Bryan Bryson, and Bonnie Berger.
\newblock Efficient integration of heterogeneous single-cell transcriptomes using scanorama.
\newblock \emph{Nat. Biotechnol.}, 37\penalty0 (6):\penalty0 685--691, June 2019.

\bibitem[Kingma \& Ba(2014)Kingma and Ba]{Kingma2014-or}
Diederik~P Kingma and Jimmy Ba.
\newblock Adam: A method for stochastic optimization.
\newblock \emph{arXiv [cs.LG]}, December 2014.

\bibitem[Kingma \& Welling(2013)Kingma and Welling]{Kingma2013-sh}
Diederik~P Kingma and Max Welling.
\newblock Auto-encoding variational bayes.
\newblock \emph{arXiv [stat.ML]}, December 2013.

\bibitem[Kiselev et~al.(2018)Kiselev, Yiu, and Hemberg]{Kiselev2018-sa}
Vladimir~Yu Kiselev, Andrew Yiu, and Martin Hemberg.
\newblock scmap: projection of single-cell {RNA}-seq data across data sets.
\newblock \emph{Nat. Methods}, 15\penalty0 (5):\penalty0 359--362, May 2018.

\bibitem[Kock et~al.(2024)Kock, Tan, Han, Ando, Jevapatarakul, Chatterjee, Lin, Buyamin, Sonthalia, Rajagopalan, Tomofuji, Sankaran, Park, Abe, Chantaraamporn, Furukawa, Ghosh, Inoue, Kojima, Kouno, Lim, Myouzen, Nguantad, Oh, Rayan, Sarkar, Suzuki, Thungsatianpun, Venkatesh, Moody, Nakano, Chen, Tian, Zhang, Tong, Tan, Tizazu, Loh, Hwang, Ho, Larbi, Ng, Won, Wright, Villani, Park, Choi, Liu, Maitra, Pithukpakorn, Suktitipat, Ishigaki, Okada, Yamamoto, Carninci, Chambers, Hon, Matangkasombut, Charoensawan, Majumder, Shin, Park, Prabhakar, and {SG10K\_Health Consortium}]{Kock2024-vi}
Kian~Hong Kock, Le~Min Tan, Kyung~Yeon Han, Yoshinari Ando, Damita Jevapatarakul, Ankita Chatterjee, Quy Xiao~Xuan Lin, Eliora~Violain Buyamin, Radhika Sonthalia, Deepa Rajagopalan, Yoshihiko Tomofuji, Shvetha Sankaran, Mi-So Park, Mai Abe, Juthamard Chantaraamporn, Seiko Furukawa, Supratim Ghosh, Gyo Inoue, Miki Kojima, Tsukasa Kouno, Jinyeong Lim, Keiko Myouzen, Sarintip Nguantad, Jin-Mi Oh, Nirmala~Arul Rayan, Sumanta Sarkar, Akari Suzuki, Narita Thungsatianpun, Prasanna~Nori Venkatesh, Jonathan Moody, Masahiro Nakano, Ziyue Chen, Chi Tian, Yuntian Zhang, Yihan Tong, Crystal T~Y Tan, Anteneh~Mehari Tizazu, Marie Loh, You~Yi Hwang, Roger~C Ho, Anis Larbi, Tze~Pin Ng, Hong-Hee Won, Fred~A Wright, Alexandra-Chloé Villani, Jong-Eun Park, Murim Choi, Boxiang Liu, Arindam Maitra, Manop Pithukpakorn, Bhoom Suktitipat, Kazuyoshi Ishigaki, Yukinori Okada, Kazuhiko Yamamoto, Piero Carninci, John~C Chambers, Chung-Chau Hon, Ponpan Matangkasombut, Varodom Charoensawan, Partha~P Majumder, Jay~W Shin, Woong-Yang Park,
  Shyam Prabhakar, and {SG10K\_Health Consortium}.
\newblock Single-cell analysis of human diversity in circulating immune cells.
\newblock \emph{bioRxiv}, pp.\  2024.06.30.601119, July 2024.

\bibitem[Korsunsky et~al.(2019)Korsunsky, Millard, Fan, Slowikowski, Zhang, Wei, Baglaenko, Brenner, Loh, and Raychaudhuri]{Korsunsky2019-lf}
Ilya Korsunsky, Nghia Millard, Jean Fan, Kamil Slowikowski, Fan Zhang, Kevin Wei, Yuriy Baglaenko, Michael Brenner, Po-Ru Loh, and Soumya Raychaudhuri.
\newblock Fast, sensitive and accurate integration of single-cell data with harmony.
\newblock \emph{Nat. Methods}, 16\penalty0 (12):\penalty0 1289--1296, December 2019.

\bibitem[Kriehuber et~al.(2001)Kriehuber, Breiteneder-Geleff, Groeger, Soleiman, Schoppmann, Stingl, Kerjaschki, and Maurer]{Kriehuber2001-em}
E~Kriehuber, S~Breiteneder-Geleff, M~Groeger, A~Soleiman, S~F Schoppmann, G~Stingl, D~Kerjaschki, and D~Maurer.
\newblock Isolation and characterization of dermal lymphatic and blood endothelial cells reveal stable and functionally specialized cell lineages.
\newblock \emph{J. Exp. Med.}, 194\penalty0 (6):\penalty0 797--808, September 2001.

\bibitem[Lopez et~al.(2018)Lopez, Regier, Cole, Jordan, and Yosef]{Lopez2018-cs}
Romain Lopez, Jeffrey Regier, Michael~B Cole, Michael~I Jordan, and Nir Yosef.
\newblock Deep generative modeling for single-cell transcriptomics.
\newblock \emph{Nat. Methods}, 15\penalty0 (12):\penalty0 1053--1058, December 2018.

\bibitem[Lotfollahi et~al.(2024)Lotfollahi, {Yuhan Hao}, Theis, and Satija]{Lotfollahi2024-vb}
Mohammad Lotfollahi, {Yuhan Hao}, Fabian~J Theis, and Rahul Satija.
\newblock The future of rapid and automated single-cell data analysis using reference mapping.
\newblock \emph{Cell}, 187\penalty0 (10):\penalty0 2343--2358, May 2024.

\bibitem[Luecken et~al.(2020)Luecken, Büttner, Chaichoompu, Danese, Interlandi, Mueller, Strobl, Zappia, Dugas, Colomé-Tatché, and Theis]{Luecken2020-bl}
Malte~D Luecken, M~Büttner, Kridsadakorn Chaichoompu, A~Danese, M~Interlandi, M~F Mueller, D~Strobl, L~Zappia, M~Dugas, M~Colomé-Tatché, and F~Theis.
\newblock Benchmarking atlas-level data integration in single-cell genomics.
\newblock \emph{Nat. Methods}, 19:\penalty0 41--50, May 2020.

\bibitem[Lähnemann et~al.(2020)Lähnemann, Köster, Szczurek, McCarthy, Hicks, Robinson, Vallejos, Campbell, Beerenwinkel, Mahfouz, Pinello, Skums, Stamatakis, Attolini, Aparicio, Baaijens, Balvert, Barbanson, Cappuccio, Corleone, Dutilh, Florescu, Guryev, Holmer, Jahn, Lobo, Keizer, Khatri, Kielbasa, Korbel, Kozlov, Kuo, Lelieveldt, Mandoiu, Marioni, Marschall, Molder, Niknejad, Raczkowska, Reinders, Ridder, Saliba, Somarakis, Stegle, Theis, Yang, Zelikovsky, McHardy, Raphael, Shah, and Schönhuth]{Lahnemann2020-kj}
David Lähnemann, Johannes Köster, Ewa Szczurek, Davis~J McCarthy, Stephanie~C Hicks, Mark~D Robinson, Catalina~A Vallejos, Kieran~R Campbell, Niko Beerenwinkel, Ahmed Mahfouz, Luca Pinello, Pavel Skums, Alexandros Stamatakis, Camille Stephan-Otto Attolini, Samuel Aparicio, Jasmijn Baaijens, Marleen Balvert, Buys~de Barbanson, Antonio Cappuccio, Giacomo Corleone, Bas~E Dutilh, Maria Florescu, Victor Guryev, Rens Holmer, Katharina Jahn, Thamar~Jessurun Lobo, Emma~M Keizer, Indu Khatri, Szymon~M Kielbasa, Jan~O Korbel, Alexey~M Kozlov, Tzu-Hao Kuo, Boudewijn P~F Lelieveldt, Ion~I Mandoiu, John~C Marioni, Tobias Marschall, Felix Molder, Amir Niknejad, Alicja Raczkowska, Marcel Reinders, Jeroen~de Ridder, Antoine-Emmanuel Saliba, Antonios Somarakis, Oliver Stegle, Fabian~J Theis, Huan Yang, Alex Zelikovsky, Alice~C McHardy, Benjamin~J Raphael, Sohrab~P Shah, and Alexander Schönhuth.
\newblock Eleven grand challenges in single-cell data science.
\newblock \emph{Genome Biol.}, 21\penalty0 (1):\penalty0 31, February 2020.

\bibitem[Pasquini et~al.(2021)Pasquini, Rojo~Arias, Schäfer, and Busskamp]{Pasquini2021-uk}
Giovanni Pasquini, Jesus~Eduardo Rojo~Arias, Patrick Schäfer, and Volker Busskamp.
\newblock Automated methods for cell type annotation on {scRNA}-seq data.
\newblock \emph{Comput. Struct. Biotechnol. J.}, 19:\penalty0 961--969, January 2021.

\bibitem[Reed et~al.(2024)Reed, Pensa, Steif, Stenning, Kunz, Porter, Hua, He, Twigger, Siu, Kania, Barrow-McGee, Goulding, Gomm, Speirs, Jones, Marioni, and Khaled]{Reed2024-ml}
Austin~D Reed, S~Pensa, Adi Steif, Jack Stenning, Daniel~J Kunz, Linsey~J Porter, Kui Hua, Peng He, Alecia-Jane Twigger, Abigail J~Q Siu, Katarzyna Kania, R~Barrow-McGee, I~Goulding, Jennifer~J Gomm, Valerie Speirs, J~L Jones, J~Marioni, and W~Khaled.
\newblock A single-cell atlas enables mapping of homeostatic cellular shifts in the adult human breast.
\newblock \emph{Nat. Genet.}, 56:\penalty0 652--662, March 2024.

\bibitem[Regev et~al.(2017)Regev, Teichmann, Lander, Amit, Benoist, Birney, Bodenmiller, Campbell, Carninci, Clatworthy, Clevers, Deplancke, Dunham, Eberwine, Eils, Enard, Farmer, Fugger, Göttgens, Hacohen, Haniffa, Hemberg, Kim, Klenerman, Kriegstein, Lein, Linnarsson, Lundberg, Lundeberg, Majumder, Marioni, Merad, Mhlanga, Nawijn, Netea, Nolan, Pe'er, Phillipakis, Ponting, Quake, Reik, Rozenblatt-Rosen, Sanes, Satija, Schumacher, Shalek, Shapiro, Sharma, Shin, Stegle, Stratton, Stubbington, Theis, Uhlen, van Oudenaarden, Wagner, Watt, Weissman, Wold, Xavier, Yosef, and {Human Cell Atlas Meeting Participants}]{Regev2017-dz}
Aviv Regev, Sarah~A Teichmann, Eric~S Lander, Ido Amit, Christophe Benoist, Ewan Birney, Bernd Bodenmiller, Peter Campbell, Piero Carninci, Menna Clatworthy, Hans Clevers, Bart Deplancke, Ian Dunham, James Eberwine, Roland Eils, Wolfgang Enard, Andrew Farmer, Lars Fugger, Berthold Göttgens, Nir Hacohen, Muzlifah Haniffa, Martin Hemberg, Seung Kim, Paul Klenerman, Arnold Kriegstein, Ed~Lein, Sten Linnarsson, Emma Lundberg, Joakim Lundeberg, Partha Majumder, John~C Marioni, Miriam Merad, Musa Mhlanga, Martijn Nawijn, Mihai Netea, Garry Nolan, Dana Pe'er, Anthony Phillipakis, Chris~P Ponting, Stephen Quake, Wolf Reik, Orit Rozenblatt-Rosen, Joshua Sanes, Rahul Satija, Ton~N Schumacher, Alex Shalek, Ehud Shapiro, Padmanee Sharma, Jay~W Shin, Oliver Stegle, Michael Stratton, Michael J~T Stubbington, Fabian~J Theis, Matthias Uhlen, Alexander van Oudenaarden, Allon Wagner, Fiona Watt, Jonathan Weissman, Barbara Wold, Ramnik Xavier, Nir Yosef, and {Human Cell Atlas Meeting Participants}.
\newblock The human cell atlas.
\newblock \emph{Elife}, 6, December 2017.

\bibitem[Rosen et~al.(2024)Rosen, Brbić, Roohani, Swanson, Li, and Leskovec]{Rosen2024-tq}
Yanay Rosen, Maria Brbić, Yusuf Roohani, Kyle Swanson, Ziang Li, and Jure Leskovec.
\newblock Toward universal cell embeddings: integrating single-cell {RNA}-seq datasets across species with {SATURN}.
\newblock \emph{Nat. Methods}, 21\penalty0 (8):\penalty0 1492--1500, August 2024.

\bibitem[Satija et~al.(2015)Satija, Farrell, Gennert, Schier, and Regev]{Satija2015-ve}
Rahul Satija, Jeffrey~A Farrell, David Gennert, Alexander~F Schier, and Aviv Regev.
\newblock Spatial reconstruction of single-cell gene expression data.
\newblock \emph{Nat. Biotechnol.}, 33\penalty0 (5):\penalty0 495--502, May 2015.

\bibitem[Shaham(2018)]{Shaham2018-kk}
Uri Shaham.
\newblock Batch effect removal via batch-free encoding.
\newblock \emph{bioRxiv}, pp.\  380816, July 2018.

\bibitem[{Tabula Sapiens Consortium}(2022)]{Tabula_Sapiens_Consortium2022-bq}
{Tabula Sapiens Consortium}.
\newblock The tabula sapiens: A multiple-organ, single-cell transcriptomic atlas of humans.
\newblock \emph{Science}, 376\penalty0 (6594):\penalty0 eabl4896, May 2022.

\bibitem[Theodoris et~al.(2023)Theodoris, Xiao, Chopra, Chaffin, Al~Sayed, Hill, Mantineo, Brydon, Zeng, Liu, and Ellinor]{Theodoris2023-qz}
Christina~V Theodoris, Ling Xiao, Anant Chopra, Mark~D Chaffin, Zeina~R Al~Sayed, Matthew~C Hill, Helene Mantineo, Elizabeth~M Brydon, Zexian Zeng, X~Shirley Liu, and Patrick~T Ellinor.
\newblock Transfer learning enables predictions in network biology.
\newblock \emph{Nature}, 618\penalty0 (7965):\penalty0 616--624, June 2023.

\bibitem[Vasighizaker et~al.(2022)Vasighizaker, Danda, and Rueda]{Vasighizaker2022-ju}
Akram Vasighizaker, Saiteja Danda, and Luis Rueda.
\newblock Discovering cell types using manifold learning and enhanced visualization of single-cell {RNA}-seq data.
\newblock \emph{Sci. Rep.}, 12\penalty0 (1):\penalty0 120, January 2022.

\bibitem[Wang et~al.(2024)Wang, Leskovec, and Regev]{Wang2024-qc}
Hanchen Wang, Jure Leskovec, and Aviv Regev.
\newblock Metric mirages in cell embeddings.
\newblock \emph{bioRxiv}, April 2024.

\bibitem[Wolf et~al.(2018)Wolf, Angerer, and Theis]{Wolf2018-qf}
F~Alexander Wolf, Philipp Angerer, and Fabian~J Theis.
\newblock {SCANPY}: large-scale single-cell gene expression data analysis.
\newblock \emph{Genome Biol.}, 19\penalty0 (1):\penalty0 15, February 2018.

\bibitem[Yang et~al.(2022)Yang, Wang, Wang, Fang, Tang, Huang, Lu, and Yao]{Yang2022-nr}
Fan Yang, Wenchuan Wang, Fang Wang, Yuan Fang, Duyu Tang, Junzhou Huang, Hui Lu, and Jianhua Yao.
\newblock {scBERT} as a large-scale pretrained deep language model for cell type annotation of single-cell {RNA}-seq data.
\newblock \emph{Nature Machine Intelligence}, 4\penalty0 (10):\penalty0 852--866, September 2022.

\end{thebibliography}
\bibliographystyle{lmrl2025_workshop}

\appendix
\section{Appendix 1: Metrics for scRNA alignment} \label{app1}

In this study, we sought to develop metrics to assess the quality of scRNA data alignments. In this Appendix, we consider the simple theoretical case of two cell types $C_1$, $C_2$, and batches $W_1$, $W_2$. Each cell in this analysis has both a cell type and a batch to which it corresponds. The probability distribution of a given cell’s type $C_1$, $C_2$ and the probability distribution of a given cell’s batch $W_1$, $W_2$  can each be approximated using a mixture of two Gaussians, with the difference between means being $\phi$ and $\omega$ respectively for cell-type and batch, i.e.,

$$ X \approx 
\begin{bmatrix}
C \\
W
\end{bmatrix} 
=
\begin{bmatrix}
N(\mu,\sigma^2) + N(\mu+\phi,\sigma^2) \\
W(\mu,\sigma^2) + W(\mu+\omega,\sigma^2) \\
\end{bmatrix} 
$$

\begin{figure}[h]

\includegraphics[width=1.0\textwidth, center]{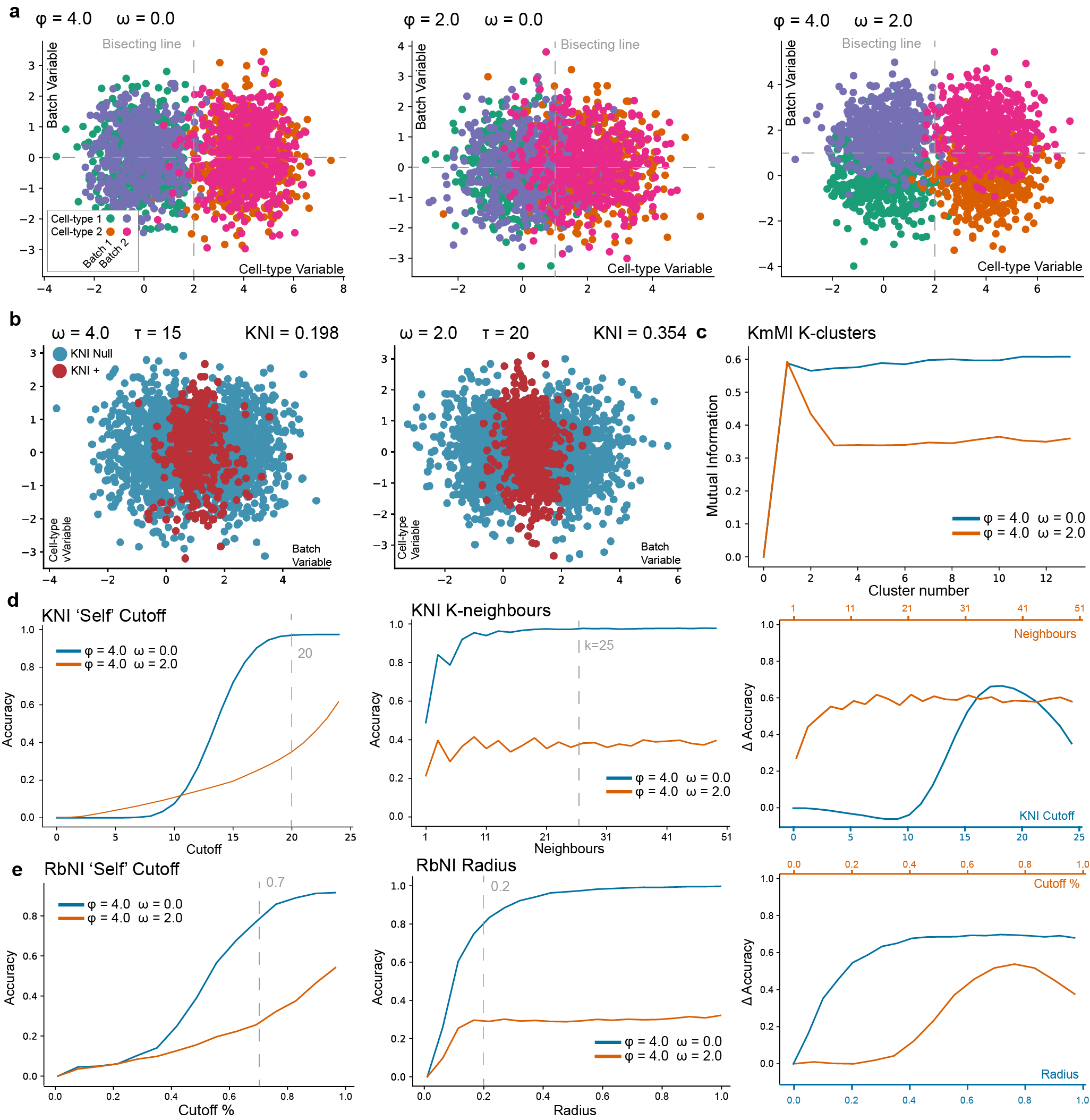}

\caption{Analysis of metric behavior on theoretical example: a) Scatter plots of the three test cases used to compare candidate metrics for evaluating scRNA alignments, the key parameters are separation of the two cell-types by $\phi$ and separation of the two batch effects by $\omega$; b) Under the KNI score, cells that are surrounded by more than $\tau$ cells from the same batch are classified as $null$, this example demonstrates the effect of this, by considering two batches separated by a batch effect $\omega=4$. Cells that are labeled as null are blue, vs. those that would be tested for label accuracy red, two regimes are considered $\tau = 15$ and $\tau = 20$; c) The KmMI score on the ideal case $\phi = 4,\omega = 0$ vs. the case of batch effects $\phi = 4,\omega = 2$  varying cluster number; f) same as (c) examples but considering the KNI score, and varying the cutoff parameter $\tau$ and number of nearest neighbors $k$, the score difference between the two test cases for the two parameters is also plotted (left); g) same as (f) but for the RbNI, where the cutoff percent $\tau*$ and radius  $r$ are varied. 
}
\label{fig:figapp1}
\end{figure}

A random sampling of a cell, $X$, is thus represented by a 2D vector where the first dimension corresponds to the cell-type and the second to the batch. In a good cell-type space, cells are well separated by biological variability $(\phi \gg 0)$, meaning a vertical line can discriminate cell types, while batch effects are small $(\omega \approx 0)$. We, therefore, are seeking a metric that is large when  $(\phi \gg 0)$,$(\omega \approx 0)$. In this example, we can use a bisecting line to demonstrate the need for explicitly penalizing batch effects in evaluation metrics (Figure \ref{fig:figapp1}a). Specifically, let's consider a classifier that splits the data along the cell-type variable line. We can see that data can be arbitrarily separated along the batch with no impact on the classifier. We thus reason that scRNA alignment metrics must explicitly penalize batch effects. Similarly, if we consider only batch-effect removal cell-clusters can be arbitrarily close together with no effect on the score. Thus, scores must factor in both batch-effect correction and preservation of cell-type groupings. We thus developed four approaches to evaluating cell-type spaces below that reward separation of cell-type groupings as defined by “ground-truth” standardized author labels while penalizing batch-effects directly. These are:
\begin{itemize}

\item Batch-corrected Silhouette score (B-Sil): To calculate this score, we calculate the Silhouette coefficient calculated on cell-type labels and deduct the coefficient calculated on batch-labels i.e., $SC(C)-SC(W)$.

\item K-means Mutual Information (KmMI): To calculate the KmMI score, we take 4 clusters K to account for all membership possibilities $x \in C_i, W_j,(i,j) \colon\{0,1\}$. We then deduct the MI between the 4 cluster labels and batch label from the MI between the cluster labels and cell-type, i.e., $MI(K;C)-MI(K;W)$. 

\item K-Neighbors Intersection (KNI): We determined that we could easily modify the K-BET score as defined in (Büttner et al. 2019), to factor in the separation of cell-types by calculating the accuracy of cross-batch prediction of cell-type. The calculation of the KNI score is as described in the main text. We termed this metric the K-neighbors intersection, as this metric calculates cell-type group membership where batches intersect in the aligned space.

\item Radius-based Neighbors Intersection (RbNI): The RbNI is equivalent to the KNI, except that: (1) The set of neighboring cells is defined by a radius, as per Radius-based Nearest Neighbors; and (2) cells with no neighbors within the radius $r$ are given an outlier label, the calculation of the RbNI score is also described in the main text.

\end{itemize}

The silhouette score has no parameters, and the KmMI has a single parameter, cluster number. Meanwhile, the KNI and RbNI metrics each have two parameters: the number of neighbors (or radius) searched ($k$, $r$) and the cutoff threshold at which to call a data point an outlier ($\tau$, $\tau*$). Changing these cutoff thresholds has the effect of increasing the size of the intersecting batch region in which cell-type labels are evaluated, as shown in Figure \ref{fig:figapp1}b. We tested how the KmMI, KNI, and RbNI metrics behave under different parameters in the theoretical cases of 'perfect' alignment $\phi=4,\omega=0$ versus a batch-effect case $\phi=4,\omega=2$ while varying metrics' parameters. Firstly, we note that the B-Sil score can distinguish these two cases, yielding a value of 0.557 for the perfect case, and 0.341 given batch effects. Secondly, we determined that the KmMI score can robustly distinguish the two cases for any cluster number greater than three (Figure \ref{fig:figapp1}c). Considering the KNI and RbNI scores, we note that both metrics are sensitive to their respective cutoff values for defining a batch effect (Figure \ref{fig:figapp1}d, e). At thresholds that separate batch effects well, these metrics appear to be comparatively insensitive to the number of cells, or radius, searched. Finally, we see that the KNI and RbNI methods show a much greater separation of score values, at optimal batch-effect cutoffs, than either of the B-Sil or KmMI methods. In this section, we propose four potential metrics for evaluating cell-type space alignment quality and show that all are capable of distinguishing an ideal theoretical cell-type space from one containing batch effects. 

\section{Appendix 2: Metric performance on simulated noise and batch-effects} \label{app2}

In a second case, we sought to understand how the scores perform on a more realistic simulated example of an aligned scRNA cell-type space $S$. To create this more realistic example we took a 30,000 cell sample from one high-quality scRNA dataset \cite{Bassez2021-ya} and embedded it into a 5-dimensional space $S \in \mathbb{R}^5$   using a basic implementation of a Variational Autoencoder with Mean Squared Error loss function \cite{Kingma2013-sh}, (VAE MSE; 2 layer encoder and decoder, 512 neurons per hidden layer, lr = 1E-4,  patience = 15). We chose this model so as to reduce bias in our model comparison, where we anticipate published models specifically designed for scRNA analysis should outperform this approach. We took the resulting embedded cell-type space and simulated both poorer alignment, and worse batch-effects, to compare how the metrics scores these spaces vs. the unmodified VAE encoding.

\begin{figure}[h]

\includegraphics[width=1.0\textwidth, center]{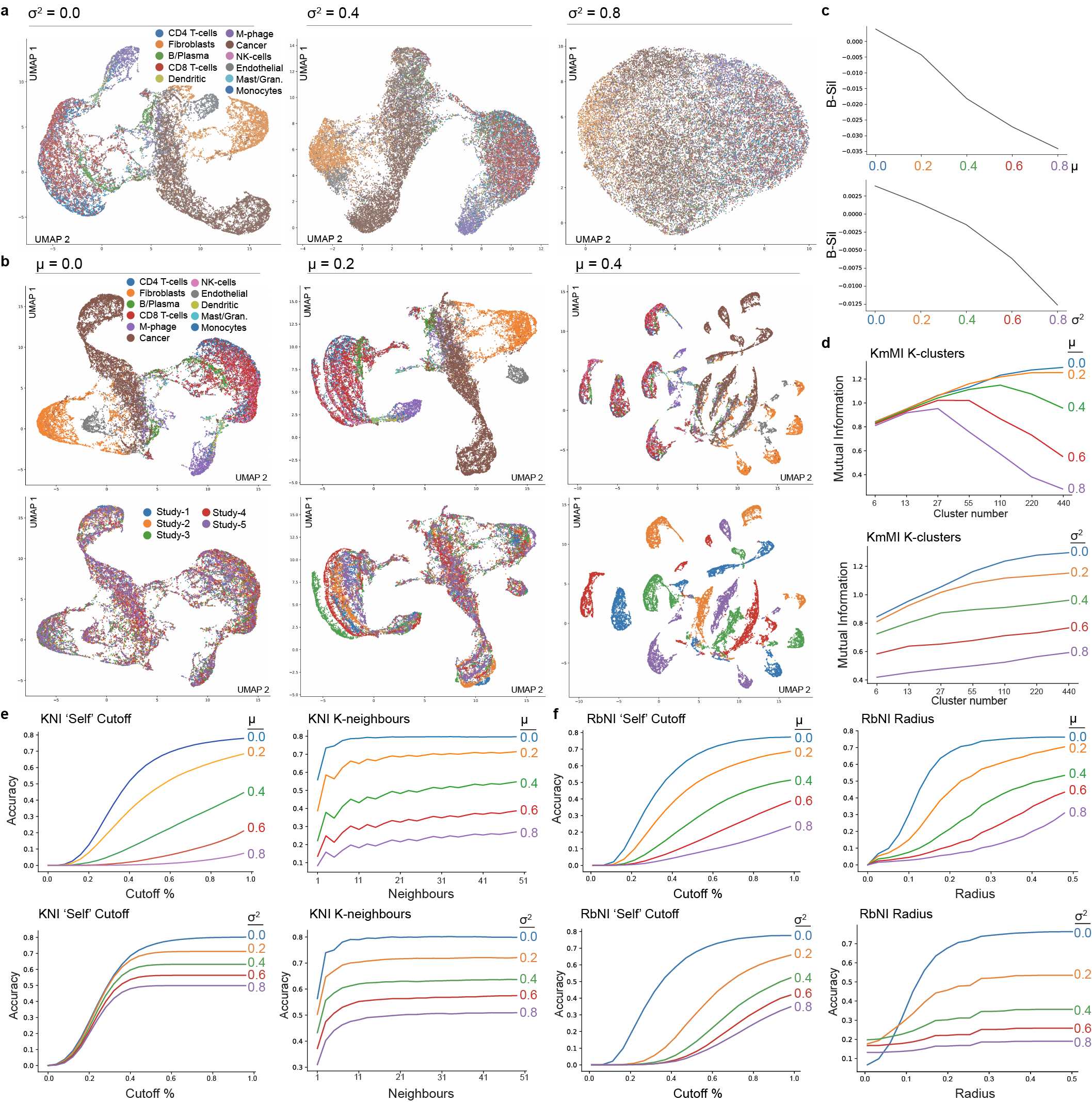}

\caption{Appendix 2: Analysis of metric behavior on real example with spiked noise and batch-effects: a) UMAP projections of 3 test cases corresponding to addition of noise to the 5 dimensional embedding space, $\sigma^2=\{0,0.4,0.8\}$,  default UMAP parameters are used as per (McInnes, Healy, and Melville 2018), cells are colored by ‘ground-truth’ cell-type; b) same as (b) but with simulation of batch-effects, where cells are split into 5 batches and  $\mu=\{0.0,0.2,0.4\}$ is added to one of the 5 dimensions for each batch. c) B-sil values for the 5 noise $\sigma^2=\{0, 0.2, 0.4, 0.6, 0.8\}$ and 5 batch test cases are shown $\mu=\{0.0,0.2,0.4,0.6, 0.8\}$; d) KmMI values for the 5 noise and batch test cases are shown as lines (colored and labeled), the parameter, cluster number is varied on the x-axis; e) same as (d), but varying the cutoff parameter and keeping Neighbors search constant, or varying the neighbor search range and keeping the cutoff constant as described in the Appendix text.
}
\label{fig:figapp2}
\end{figure}

To simulate progressively worse alignment, we introduced random Gaussian noise to the cell-type space $S + N(\mu,\sigma^2 )$   where $\mu=0$ and $\sigma^2=\{0,0.2,0.4,0.6,0.8\}$. Data was then re-normalized to unit mean and variance after the addition of noise. UMAP projections of the cell-type spaces for cases $\sigma^2=\{0,0.4,0.8\}$ are given in Figure \ref{fig:figapp2}a and highlight the progressive increase in cell-type cluster overlap associated with introduction of this noise. To simulate progressively worse batch effects, we split the embedded data into 5 equally sized groups $C_i,\ i=\{1\dots5\}, C \in \mathbb{R}^5$ and add a constant value $\mu$ to the respective dimension for each group where $\mu=\{ 0,0.1,0.2, 0.3,0.4\}$. Data was then re-normalized to unit mean and variance after the addition.  The effect of this can be seen in UMAP projections of the embedding space for $\mu=\{0.0,0.2,0.4\}$ Figure \ref{fig:figapp2}b. 

We then evaluated metric performance on these test cases, to understand how well each metric could distinguish noise and batch-effects: 

\begin{itemize}

\item Batch Adjusted Silhouette Score (B-sil): In both the cases of simulated noise and simulated batch-effect, the B-sil decreased linearly with increasing noise, or batch effect (Figure \ref{fig:figapp2}c). Of note, however, these values were dramatically smaller than those calculated for the theoretical case of two batch effects and two cell types, likely a result of the higher dimensionality of the embedding space and many more clusters. The B-sil score shows such variability based on the underlying dataset, thus making it a potentially poor choice for evaluating model performance on benchmarks, especially as complexity increases. Moreover, changes in the score in response to the addition of significant batch effects were 3x smaller than those seen in response to the addition of noise, another potential issue with using this score as a metric of alignment quality.

\item K-means Mutual Information (KmMI): The KmMI score showed a linear relationship to the addition of noise and was only able to separate the smallest addition of noise $\sigma^2=\{0.2\}$ at cluster numbers over 100 (Figure \ref{fig:figapp2}d). However, at these higher cluster numbers, robust separation between all scenarios was observed. The KmMI metric showed even greater insensitivity to the addition of small batch effects than it did to noise. Here, greater than 440 clusters were needed to even begin to see a separation between $\mu=0.0$ and $\mu=0.1$; at this number of clusters, only 70 cells on average are present per cluster. This is a number of cells to those analyzed in the local neighborhood by the Radius-based and K-neighbors metrics. This analysis indicates the KmMI as a poor metric for identifying batch effects and suggests local analysis of cell-type space is likely better.  

\item K-Neighbors intersection (KNI): 	The KNI score showed a decrease in score for all cutoff values greater than $\tau \approx 0.5$, indicating this score is effective over a reasonably wide set of parameters (Figure \ref{fig:figapp2}e). Of note, this is the same value of $\tau$ for which optimal separation of scores was seen in the theoretical case described in the first section, indicating this parameter may also be robust to dimensionality and dataset complexity. Varying the number of neighbors k while setting $\tau = 0.8$ of the value of $k$  also highlights that for values of k > 10, a robust separation between all scenarios is observed. The KNI also emerges as being very effective in identifying and quantifying the addition of batch effects to the aligned cell-type space  (Figure \ref{fig:figapp2}e). Varying the cutoff number $\tau$ shows that all batch effects can be well separated for values of $\tau>0.5$ (k set at 25). Setting $\tau = 0.8$ and varying the number of neighbors k used for label calling demonstrates that for $k>5$ there is good separation between all batch effect scenarios, again the best separation is seen for similar cutoff values in this data, as is seen in the simple theoretical case given in the first section, indicating limited sensitivity in the cutoff value as the dimensionality and complexity of the data increases. The KNI overall emerges as a metric that performs well at identifying simulated batch-effects and noise.

\item Radius-based Neighbors intersection (RbNI): 	the RbNI score showed a similar nonlinear sensitivity to the RbNN score with respect to the addition of noise, a valuable property as noted above. Similar to the KNI, the RbNI score was insensitive to the selection of the cutoff percent $\tau*$, for $\tau*>0.5$ when the radius was fixed at a value of 0.3, indicating stability here (Figure \ref{fig:figapp2}e). This result also matched that seen in the theoretical test case above, suggesting that this parameter is likely stable of cell-type space dimensionality and complexity. Finally, a similar relationship between the selection of r and sensitivity to the addition of noise was seen between the RbNI and RbNN methods when a cutoff of $\tau* = 0.75$ was used. Again, we find that the RbNI metric performs well at identifying both batch-effects and noise.

Based on this analysis, we find the KNI and RbNI scores perform particularly well at identifying both batch effects and noise and appear to generalize well between datasets. We focus on and recommend the KNI score, given its similarity to the K-bet score that is now well-described and used for batch-effect detection, though report RbNI values to demonstrate that both metrics can be used and correlate well, indicating insensitivity to the method of neighborhood search.
\end{itemize}

\section{Appendix 3: Dataset Construction} \label{app3}

Raw scRNA UMI count matrices were obtained from public repositories. Quality control followed the original author filters. Cells labeled by the authors as; (i) Unknown; (ii) Undetermined; or (iii) Mixed were excluded from benchmark analysis. Gene identifiers were standardized across studies based on (i) Human Protein Atlas (HPA) versions 13 to 20; and (ii) ENSEMBL GRCh38 versions 78 to 103. Priority was given to HPA identifiers. For scMARK, genes present in all datasets were used for training. For scREF, genes common to 30 datasets or more were used for training. In cases where the authors provided only general T-cell annotations, we used Azimuth’s Human PBMC signatures (Hao et al. 2021)  to assign those cells into CD8+ or CD4+ cells. Gamma-delta T-cells were also included for scREF. The studies present in scMARK are given in Appendix Table A, while the studies given in scREF are provided in Table Appendix Table B. Mappings of author labels to standardized cell-type labels and specific gene details can be found on github [LINK PROVIDED AFTER BLINDED REVIEW].

\begin{table}[]
\begin{tabular}{| l | l | l | p{25mm} | p{25mm} | p{30mm} |}
 \hline
 \multicolumn{6}{|c|}{Appendix Table A scMARK Datasets} \\
 \hline
First   Author & Year & PMID     & Normal Tissues                & Cancer Tissues                    & Technology                        \\ \hline
Azizi          & 2018 & 29961579 & Breast                        & Breast                            & InDrop                            \\
Bassez         & 2021 & 33958794 & None                          & Breast                            & \RaggedRight 10x Ch 5' or 3'     \\
Bi             & 2021 & 33711272 & None                          & Kidney                            & \RaggedRight 10x Ch. 3' v2            \\
Elyada         & 2019 & 31197017 & Pancreas                      & Pancreas                          & \RaggedRight 10x Ch. 3' v2            \\
Karlsson       & 2021 & 34321199 & \RaggedRight Breast, Kidney, Lung, Ovarian & None                              & \RaggedRight 10x Ch. 5' or 3'     \\
Lee            & 2020 & 32451460 & Colorectal                    & Colorectal                        & \RaggedRight 10x Ch. 3prime v2            \\
Nath           & 2021 & 34031395 & None                          & Ovarian                           & \RaggedRight 10x Ch. 3' v3 and iCell8 \\
Peng           & 2019 & 31273297 & Pancreas                      & Pancreas                          & \RaggedRight 10x Ch. 3' v2            \\
Qian           & 2020 & 32561858 & \RaggedRight Colorectal, Lung, Ovarian     & \RaggedRight Breast, Colorectal, Lung, Ovarian & 10x Ch. 5' or 3' v2  \\
Slyper         & 2020 & 32405060 & None                          & Lung                              & \RaggedRight 10x Ch. 3' v2 or v3      \\
Zhang          & 2021 & 34099557 & Kidney                        & Kidney                            & \RaggedRight 10x Ch. 3' v2           
\\ \hline     
\end{tabular}
\label{table:2}
\end{table}

\begin{table}[]
\begin{tabular}{| l | l | l | p{40mm} |p{40mm} |}
 \hline
 \multicolumn{5}{|c|}{Appendix Table B scREF Datasets} \\
 \hline
 
First   Author & Year & PMID          & Normal Tissues                    & Technology                             \\ \hline
Adams          & 2020 & PMID:32832599 & Lung                              & 10x Ch. 3' v2                          \\
Aida           & 2023 & Cell x Gene            & Blood                             & 10x Ch. 5' v2                          \\
Andrews        & 2022 & PMID:34792289 & Liver                             & 10x Ch. 3' v2 or v3                    \\
Bakken         & 2021 & PMID:34616062 & Brain                             & 10x Ch. 3' v3                          \\
Bautista       & 2021 & PMID:33597545 & Thymus                            & 10x Ch. 3' v2 or v3                    \\
Bhatnakshatri  & 2021 & PMID:33763657 & Breast                            & 10x Ch. 3' v3                          \\
Cillo          & 2020 & PMID:31924475 & Blood, Head-and-neck              & 10x Ch. 3' v2                          \\
Demicheli      & 2020 & PMID:32624006 & Skeletal-muscle                   & 10x Ch. 3' v2                          \\
Elmentaite     & 2021 & PMID:34497389 & \RaggedRight Colorectal, Intestine, Lymph-node & 10x Ch. 5' v2 or 3' v2 \\
Fan            & 2019 & PMID:31320652 & Ovarian                           & 10x Ch. 3' v2                          \\
Garciaalonso   & 2021 & PMID:34857954 & Uterus                            & 10x Ch. 3' v2 or v3                    \\
Guo            & 2018 & PMID:30315278 & Testis                            & 10x Ch. 3' v2                          \\
Habermann      & 2020 & PMID:32832598 & Lung                              & 10x Ch. 3' v2 or 5'                    \\
Han            & 2020 & PMID:32214235 & Many / Whole Organism             & microwell-seq                          \\
He             & 2020 & PMID:33287869 & Many / Whole Organism             & 10x Ch. 5'                    \\
Henry          & 2018 & PMID:30566875 & Prostate                          & 10x Ch. 3' v2                          \\
Hildreth       & 2021 & PMID:33907320 & Adipose                           & 10x Ch. 3' v3                          \\
Jones          & 2022 & PMID:35549404 & Many / Whole Organism             & 10x Ch. 3' v2 or 5' v2                 \\
Kfoury         & 2021 & PMID:34719426 & Bone-marrow                       & 10x Ch. 3' v2                          \\
Kong           & 2023 & PMID:36720220 & Colorectal, Intestine             & 10x Ch. 3' v2 or v3                    \\
Lake           & 2023 & PMID:37468583 & Kidney                            & 10x Ch. 3' v3                          \\
Lein           & 2023 & Cell x Gene            & Brain                             & \RaggedRight 10x Ch. 3' v3 or 10x multiome          \\
Lengyel        & 2022 & PMID:36543131 & Ovarian                           & 10x Ch. 3' v3 or Drop-seq              \\
Liang          & 2023 & PMID:37388908 & Eye                               & 10x Ch. 3' v3                          \\
Litvinukova    & 2020 & PMID:32971526 & Heart                             & 10x Ch. 3' v3 or 10x Ch. 3' v2         \\
Lukassen       & 2020 & PMID:32246845 & Bronchus                          & 10x Ch. 3' v2                          \\
Macparland     & 2018 & PMID:30348985 & Liver                             & 10x Ch. 3' v2                          \\
Madissoon      & 2019 & PMID:31892341 & \RaggedRight Head-and-neck, Lung, Spleen       & 10x Ch. 3' v2                          \\
Mayr           & 2021 & PMID:33650774 & Lung                              & Drop-seq                               \\
Menon          & 2019 & PMID:31653841 & Eye                               & 10x Ch. 3' v3 and Seq-Well             \\
Nie            & 2022 & PMID:35504286 & Testis                            & 10x Ch. 3' v3                          \\
Nowickiosuch   & 2023 & PMID:36929873 & \RaggedRight Gastric, Head-and-neck, Intestine & 10x Ch. 3' v2 or v3                    \\
Pal            & 2021 & PMID:33950524 & Breast                            & 10x Ch. 3'                             \\
Parikh         & 2019 & PMID:30814735 & Colorectal                        & \RaggedRight 10x Ch. 5' v2 and Smart-seq2           \\
Perez          & 2022 & PMID:35389781 & Blood                             & 10x Ch. 3' v2                          \\
Qadir          & 2020 & PMID:32354994 & Pancreas                          & 10x Ch. 3' v2                          \\
Reed           & 2023 & PMID:38548988 & Breast                            & 10x Ch. 3' v3                          \\
Siletti        & 2023 & PMID:37824663 & Brain                             & 10x Ch. 3' v3                          \\
Sohni          & 2019 & PMID:30726734 & Testis                            & 10x Ch. 3' v2                          \\
Soleboldo      & 2020 & PMID:32327715 & Skin                              & 10x Ch. 3' v2                          \\
Vangalen       & 2019 & PMID:30827681 & Bone-marrow                       & Seq-Well                               \\
Ventotormo     & 2018 & PMID:30429548 & Blood, Decidua, Placenta          & \RaggedRight 10x Ch. 3' v2 and Smart-seq2           \\
Wang           & 2020 & PMID:31915373 & Heart                             & SMARTScribe/Takara                     \\
Wang           & 2020 & PMID:31753849 & Colorectal, Intestine, Heart      & 10x Ch. 3' v2                          \\
Wiedemann      & 2023 & PMID:36732947 & Skin                              & 10x Ch. 3' v2 or v3                    \\
Zhao           & 2020 & PMID:33173058 & Testis                            & 10x Ch. 3' v2
\\ \hline               
\end{tabular}
\label{table:3}
\end{table}

\section{Appendix 4: Model Methods} \label{app4}

Cell-type space alignment parameters and methods were implemented as follows:

\begin{itemize}
\item \textbf{PCA:} Highly variable genes were selected based on higher dispersion than genes with similar mean expression \cite{Satija2015-ve}, implemented in scanpy v1.7.2 \cite{Wolf2018-qf}. PCA was run on scaled, normalized expression of highly variable genes.

\item \textbf{RPCA:} Implemented in R using Seurat (v4.0.3) \cite{Hao2021-zp} with top-10 larger samples as references for anchor detection (parameters: dims=10, npcs=10, k.filter=150, k.weight=100). Output from RunPCA (npcs=10) and RunUMAP (n.components=10) with assay="SCT" were used for KNI/RbNI calculations.

\item \textbf{Harmony:} PCs identified from highly variable genes with PCA were passed to harmony-pytorch v.0.1.7, using default parameters.

\item \textbf{scVI 2L Sample:} Reimplemented the scVI variational auto-encoder \cite{Lopez2018-cs} with sample level batch-correction. 1) Used 2-layer encoder and decoders. 2) 512 hidden nodes per linear layer. 3) Dropout regularization (0.1 probability). 4) Batch normalization between hidden layers. 5) ReLU activation function. 6) 10-dimensional latent space with Normal distribution. 7) Zero-Inflated Negative Binomial distribution for gene counts. 8) Adam optimizer (learning rate = 1E-4, weight decay = 1E-5, eps = 0.01). 9) Early stopping (patience = 15 epochs). 10) Batch size 64, maximum 100 epochs. 11) Implemented in Python using Pytorch (1.7.0). 12) One-hot Batch ID vectors for unique Sample ID (386 batches/samples across 10 studies).

\item \textbf{scVI 4L Sample:} Optimized scVI model with 4 layers in encoder/decoder and patience = 5 for early stopping.

\item \textbf{scVI* (ScVI 4L-NoL-NoB Both):} Optimized scVI model without batch ID requirement in encoder. 1) Removed explicit library size handling. 2) No batch ID vector injection into encoder layer. 3) Two-hot batch ID vector encoding Sample ID (386) and study ID (11). 4) Learning rate = 5E-5.

\item \textbf{scGPT:} Used authors' tutorials for zero-shot and fine-tuned embeddings (accessed March 25th, 2024). 512-dimensional embeddings from fine-tuned models reduced to 10 dimensions using two-layer autoencoder with cosine similarity loss.

\item \textbf{geneFormer:} Used authors' zero-shot pipeline for preprocessing, tokenization, and embedding (accessed March 27th, 2024).
\end{itemize}

For geneFormer and scGPT, dataloaders used TileDB database, while VAE models (scVI, BA-scVI) loaded data directly from H5AD files. All dataloaders and model training procedures leveraged PyTorch lightning library.

For scVI, PCAscmap, Harmony, and BA-scVI, the count matrices were normalized on a per-cell basis using Scanpy v1.7.2 \cite{Wolf2018-qf}, by dividing each cell by its total count over all genes. The normalized count was then multiplied by a scale factor of 10,000, after which a log(X+1) transformation was applied. For RPCA, Seurat’s SCTransform normalization was used with default parameters \cite{Hao2021-zp}.

\subsection{BA-scVI Architecture}
Batch-Adversarial scVI (BA-scVI) leverages the same core architecture as scVI, but makes use of an adversarial framework for removing batch effects. The key difference is where scVI injects one-hot batch ID vectors into the encoder and decoder layers, BA-ScVI takes an adversarial learning approach to learning and removing batch-effects. 

Here discriminators seek to predict the batch-ID $b_i$ using the encoder outputs and decoder inputs. Namely, the discriminator $D$ seeks to minimize loss $L$ with respect to batch-ID on the encoder outputs $W_E$ and decoder outputs $W_D$. The encoder and decoder weights are frozen in this step. We use cross entropy loss such that:

$$L_{disc.} = \sum_{i} b_i \log(p_i) + \sum_{i} b_i \log(q_i), \quad p_i = D(W_E), \quad q_i = D(W_D)$$

The inference network then seeks to: 1) Maximize the probability of the posterior, which in this case we use a Zero-Inflated Negative Binomial (ZINB) distribution as per \cite{Lopez2018-cs}; 2) Minimize KL-divergence of the embedding distribution $z$ and library encoder $l$ \cite{Kingma2013-sh}; and 3) Maximize discriminator loss, i.e.

$$L_{BAscVI} = -\mathbb{E}_{q(z,l|x)} \log p(x|z,l) + D_{KL}(z) + D_{KL}(l) - \beta L_{disc.}$$

The discriminator and inference networks are then trained in sequential steps with the first step used to update weights on the discriminator networks and the second step weights on the inference network. An optimal regimen for training was identified (Table S3) that leveraged an Adam optimizer \cite{Kingma2014-or}, with learning rate = 5E-5 for the inference network, 1E-2 for the discriminator network; weight decay = 1E-5; and eps = 0.01, with a batch-size 64 and for a maximum of 100 epochs; $\beta = 1000$ was used for the model evaluated in the main text. Values of $\beta = 10^{(-1 .. 5)}$ were tested and an optimal value chosen. In this optimal training regime a two-hot batch ID vector was also used that encoded 'Both' Sample ID (386 long), and study ID (11 long) was also used.

\subsection{Model Training Details}
Models were trained on scREF, the scREF/scREF-mu atlas using a regime optimized on a smaller benchmark scMARK that we discuss in the supplemental, with the exception of our handling of a standardized gene set for training. For scMARK genes common to all datasets were used. For scREF and the joint atlas we took a list of genes common across 30 datasets or more. To handle missing genes for a specific dataset, we then applied a mask to the reconstruction loss function at train time, such that only genes present in the dataset affected the overall loss. This mask was not applied to either the encoder or decoder, and thus will not affect prediction results. For the joint atlas, we used ENSEMBL v110. On scREF-mu, mouse genes identifiers common to all datasets were used (Table S1).

\section{Appendix 5: KNI scores on new cell-type labels} \label{app5}

\begin{figure}[h]

\includegraphics[width=0.8\textwidth, center]{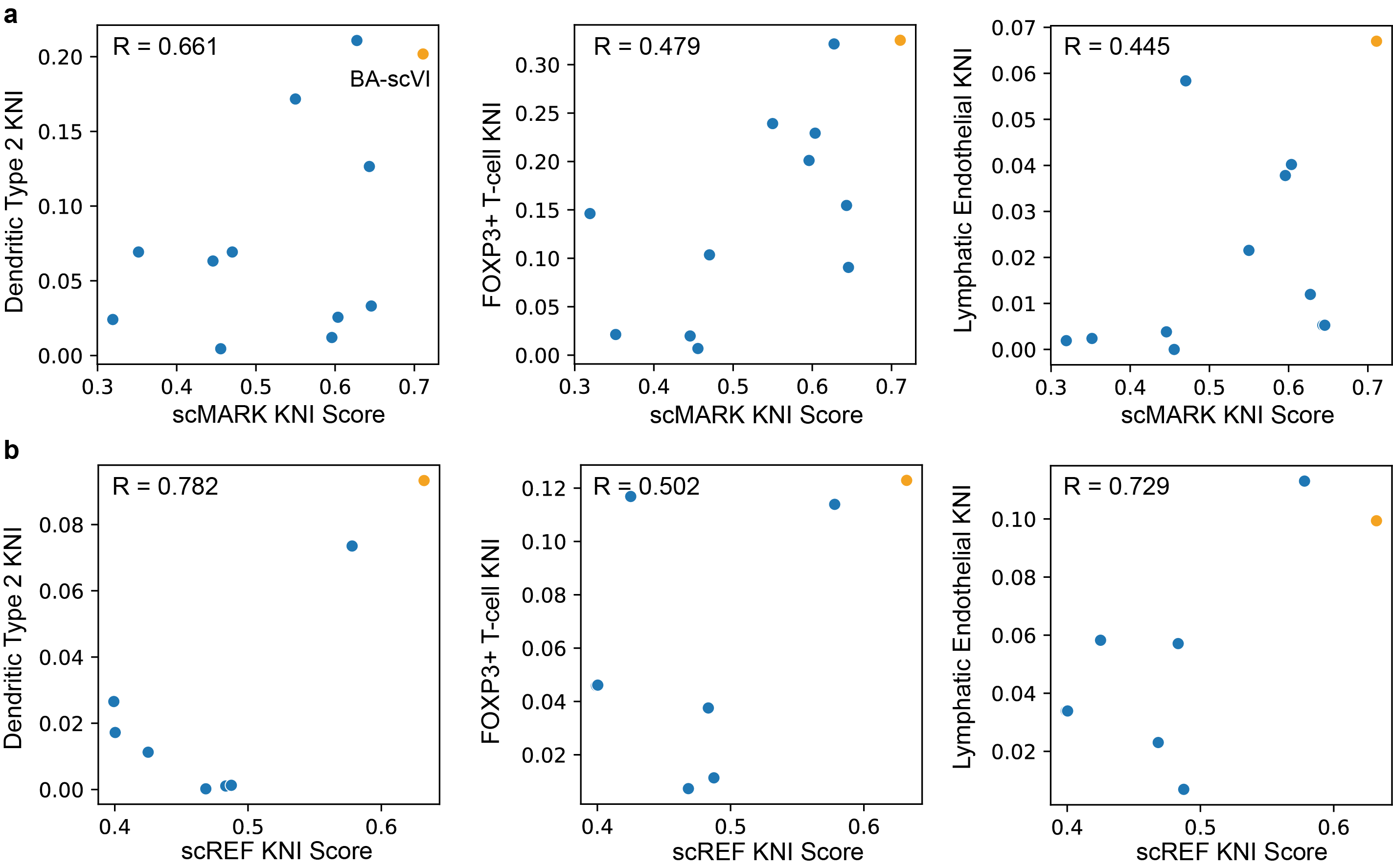}

\caption{Correlation of the KNI score between standardized and newly defined cell-types: a) Scatter plots show the correlation between KNI scores achieved on the scMARK dataset using standardized author labels (x-axis) and three cell-types (y-axis) defined by non-zero gene expression of CLEC9A (Dendritic-cell subtype), FOXP3 (T-regs), and CCL21 (Lymphatic Endothelial) in the scMARK dataset; b) the same as (a), but comparing KNI scores obtained on the scREF dataset and KNI scores obtained on the three cell-types defined in the scREF dataset}
\label{fig:app5}
\end{figure}

A key assumption in our benchmarking approach is that consensus author labels can be used to identify models that align scRNA data effectively, and that the KNI readouts capture this. To test this assumption, we evaluated model performance using the KNI score on three well-defined cell types not included in the original dataset and that we define by gene expression vs. author labels. Specifically, we use (1) CLEC9A+ Dendritic cells \cite{Caminschi2008-ds};   (2) T regulatory T-cells (T-regs), expressing FOXP3 \cite{Fontenot2003-og}; (3) and Lymphatic Endothelial cells, positive for CCL21 \cite{Kriehuber2001-em}. We assigned new cell-type labels to these cells based on the non-zero expression of the respective marker gene and then assessed the model's ability to identify these cell types based on the KNI score. We note that only a subset of the true set of these cells is likely labeled by this approach due to  high dropout rates in scRNA-seq. Across all three cell types, we noted that positive correlations were seen between the newly defined cell types and both the KNI scores on scMARK (Figure \ref{fig:app5}a) and scREF (Figure \ref{fig:app5}b. Overall, this indicates that methods that score well under the KNI on standardized labels are also best for identifying new-cell type groupings as defined by gene expression. Importantly, BA-scVI, the top performing model, was the best performing model in 4/6 tests, and second from the top in the remaining two. Overall, this analysis thus supports our initial assumption that author labels approximate a ground truth, can be used for effectively assessing model performance, and supports the validity of the scMARK and scREF alignments presented here.

\end{document}